\documentclass[preprint,12pt]{elsarticle}

\usepackage{amssymb}
\usepackage{amsmath}
\usepackage{amsthm}
\usepackage{booktabs}    
\usepackage{float}      
\usepackage{array} 
\usepackage{url}

\journal{Journal of Computational Physics}

\usepackage{xcolor}
\newif\ifcomments
\commentstrue  

\usepackage[normalem]{ulem}

\begin{document}

\begin{frontmatter}

\title{PTL-PINNs: Perturbation-Guided Transfer Learning with
Physics-Informed Neural Networks for Nonlinear Systems}

\affiliation[label1]{organization={Department of Earth Science and Engineering, Imperial College},
            addressline={Exhibition Road}, 
            city={London},
            postcode={SW7 2AZ}, 
            country={England}}

\fntext[fn1]{Corresponding author email: \texttt{d.alexandrino2010@gmail.com}}

\author[label1]{Duarte Alexandrino \fnref{fn1}}

\author[label1]{Ben Moseley}

\affiliation[label2]{organization={Harvard John A. Paulson School Of Engineering And Applied Sciences},
            addressline={150 Western Ave}, 
            city={Boston},
            postcode={MA 02134}, 
            country={United States}}

\author[label2]{Pavlos Protopapas}

\begin{abstract}
Accurately and efficiently solving nonlinear differential equations is crucial for modeling dynamic behavior across science and engineering. Physics-Informed Neural Networks (PINNs) have emerged as a powerful solution that embeds physical laws in training by enforcing equation residuals. However, these struggle to model nonlinear dynamics, suffering from limited generalization across problems and long training times. To address these limitations, we propose a perturbation-guided transfer learning framework for PINNs (PTL-PINN), which integrates perturbation theory with transfer learning to efficiently solve nonlinear equations. Unlike gradient-based transfer learning, PTL-PINNs solve an approximate linear perturbative system using closed-form expressions, enabling rapid generalization with the time complexity of matrix--vector multiplication. We show that PTL-PINNs achieve accuracy comparable to various Runge--Kutta methods, with computational speeds up to one order of magnitude faster. To benchmark performance, we solve a broad set of problems, including nonlinear oscillators across various damping regimes, the equilibrium-centered Lotka--Volterra system, the KPP--Fisher and the Wave equation. Since perturbation theory sets the accuracy bound of PTL-PINNs, we systematically evaluate its practical applicability. This work connects long-standing perturbation methods with PINNs, demonstrating how perturbation theory can guide foundational models to solve nonlinear systems with speeds comparable to those of classical solvers.
\end{abstract}

\begin{graphicalabstract}
\centering
\includegraphics[width=\textwidth]{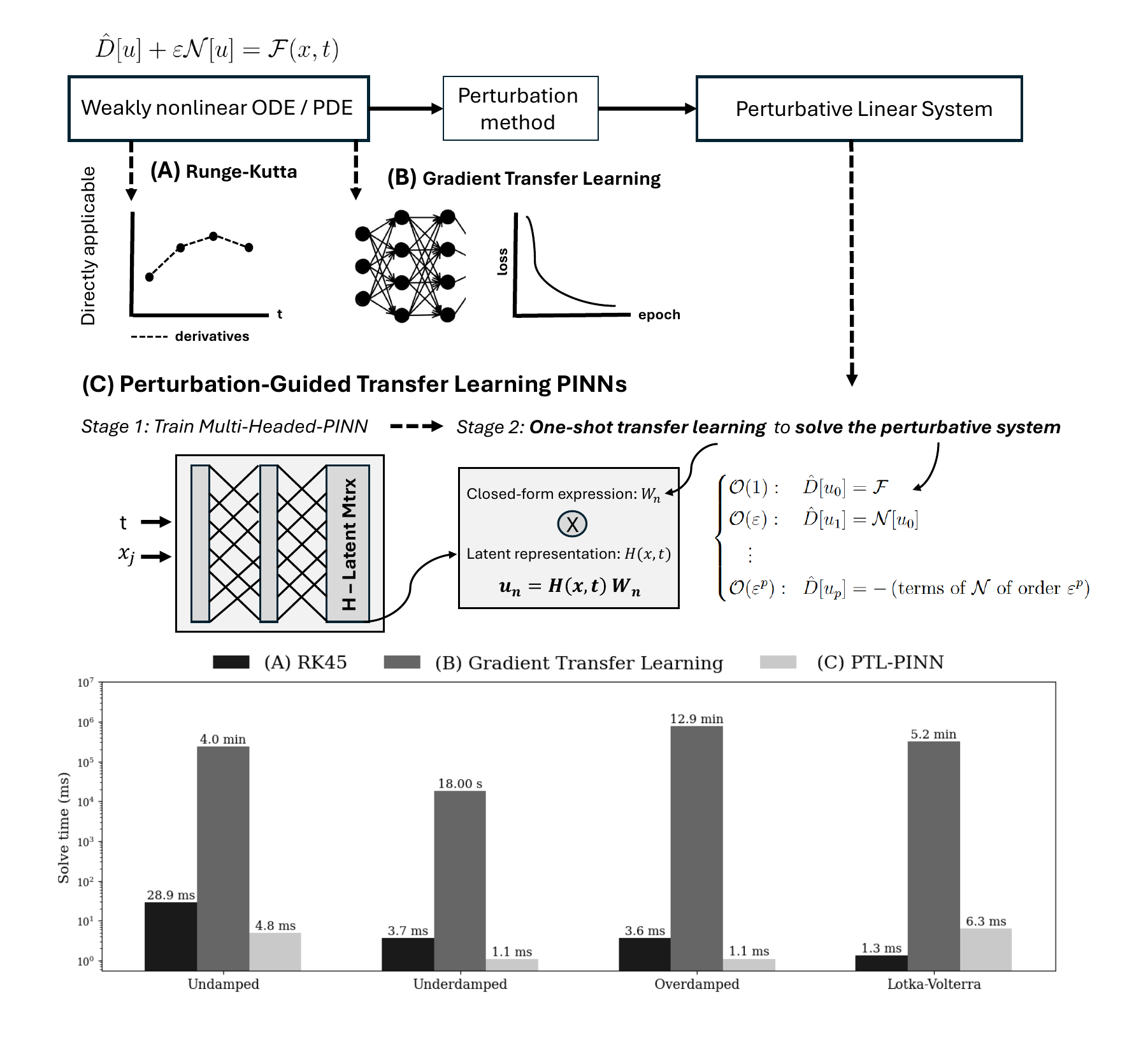}
\end{graphicalabstract}

\begin{highlights}

    \item We introduce a Perturbation-Guided Transfer Learning framework for PINNs (PTL-PINNs), which combines perturbation theory with one-shot transfer learning to efficiently solve the sequence of linear systems that approximate weakly nonlinear ODEs and PDEs.
    
    \item We benchmark PTL-PINNs across a broad class of weakly nonlinear systems, including second-order and coupled ODEs, as well as PDEs that are first-order in time and first-order or second-order in space.
    
    \item We show that PTL-PINNs can solve weakly nonlinear systems up to an order of magnitude faster than Runge--Kutta methods by generalizing to new equations via closed-form updates of the output weights, avoiding gradient-based optimization.
    
    \item We derive general $n$th-order perturbation equations and forcing functions using both standard and Lindstedt--Poincar\'e methods. These derivations enable PTL-PINNs to solve perturbative systems to higher orders and guide the selection of pretraining equations.
    
    \item We demonstrate how perturbation methods can be systematically integrated with PINNs. We show that PTL-PINNs reproduce known perturbation-theory failure modes and that the number of perturbation basis functions scales linearly with runtime and logarithmically with error.

\end{highlights}


\end{frontmatter}



\section{Introduction}

Differential equations are ubiquitous in modeling real-world phenomena. A wide range of scientific and engineering problems can be reduced to formulating and solving differential equations. However, most nonlinear differential equations do not admit analytical solutions.
The standard approach to solving them is numerical methods such as the finite difference, finite element or finite volume method \cite{larsson2003}. These provide robust approximations but require iterative time-stepping schemes, needing to be rerun for every new equation. This makes large-scale simulations computationally expensive and limits their practicality in settings that demand repeated evaluations of nonlinear differential equations, such as design optimization or Bayesian inference \cite{martins2021engineering, jameson1988aerodynamic}.

There has been a growing trend towards using neural networks to accelerate simulations in recent years. These have been used to accelerate and improve the results of numerical solvers \cite{kochkov2021machine, bar2019learning, stevens2020enhancement}. Moreover, they have been used to entirely replace them through architectures such as Neural Operators \cite{kovachki2023neural} and Physics-Informed Neural Networks (PINNs) \cite{lagaris1998artificial, raissi2019physics}. While Neural Operators learn the mapping between input and solution function spaces purely from data \cite{lu2021learning, li2020fourier, matveev2025lightweightdiffusionmultiplieruncertainty}, PINNs leverage automatic differentiation to enforce the residuals of the governing differential equations directly in the loss function, making them particularly effective in low-data regimes. In this process, they learn the functional representations of equations, offering rapid forward pass evaluations \cite{karniadakis2021physics}.

PINNs have been successfully applied to solve nonlinear differential equations across various areas, including fluid dynamics, wave propagation and chaotic systems \cite{zhang2020physics, uddin2023wavelets, moseley2020solving}. Despite being good solvers for relatively trivial problems, they easily fail to learn relevant physical phenomena for even slightly more complex equations. For nonlinear oscillatory systems, such as those considered in this work, PINNs tend to learn higher frequencies more slowly than lower frequencies, which is known as the spectral bias \cite{moseley2023finite}. The failure modes of PINNs are often not due to the lack of expressivity in the neural network architecture, but rather to a loss landscape being highly non-convex and hard to optimize \cite{krishnapriyan2021characterizing}. More fundamentally, the practical applicability of PINNs is limited by poor generalizability across equations. In the standard approach, a PINN is trained to minimize the residual of a specific differential equation and its associated initial/boundary conditions. Consequently, the learned weights fail to generalize when physical parameters and initial/boundary conditions change \cite{wong2023baldwin}.

Transfer learning has been used to address this limitation \cite{wang2025transfer, goswami2020transfer, yuan2022pinn}. This method involves reusing a pretrained neural network to infer the solution for a different equation. In practice, it is performed by initializing a new network with the weights of the pretrained PINN and optimizing the weights of the last layer(s) based on the loss function of a new and closely related problem. Since the network's weights already have encoded the general features of the underlying physics, this requires significantly fewer iterations than training a new model. This process still relies on gradient-based optimization to update model weights, which remains orders of magnitude slower than classical numerical solvers.

A faster alternative is one-shot transfer learning \cite{desai2024oneshot}. This approach leverages a Multi-Headed-PINN architecture, which approximates equations of similar form with a shared latent representation \cite{zou2023lhydramultiheadphysicsinformedneural}. In one-shot transfer learning, the solution to a new equation is obtained from a closed-form expression rather than through gradient descent, reducing inference to matrix operations. However, one-shot transfer learning is only applicable to linear differential equations, since their loss functions are convex with respect to the weights. Recent work with PINNs has begun to explore perturbative strategies to relax this linearity restriction by approximating nonlinear problems through structured linear expansions \cite{lei2023oneshottransferlearningnonlinear, auroy2025oneshottransferlearningnonlinear}. These approaches motivate the idea that weakly nonlinear systems may be tractable via linear surrogate problems, but they do not yet provide a general framework that combines perturbation theory with reusable latent representations and closed-form inference.

In this work, we introduce \emph{Perturbation-Guided Transfer Learning for Physics-Informed Neural Networks} (PTL-PINNs), a general framework that extends one-shot transfer learning to weakly nonlinear differential equations. The central idea is to reformulate a nonlinear problem as a hierarchy of linear differential equations derived from perturbation theory \cite{bender1999advanced} and to solve each linear subproblem via one-shot transfer learning using a shared latent representation learned by a Multi-Headed-PINN. While perturbative one-shot approaches have been explored for specific systems, PTL-PINNs provide a unified and reusable framework that applies across ordinary, coupled and partial differential equations. The framework is designed to systematically incorporate different perturbative formulations, which allow known limitations of perturbation theory, particularly in oscillatory systems, to be addressed within a common computational setting. By replacing gradient-based optimization with closed-form updates of the output weights at each perturbation order, PTL-PINNs enable rapid generalization to new equations, achieving runtimes competitive with classical numerical solvers from the Runge--Kuta family \cite{virtanen2020scipy}.

Various formulations of perturbation methods have been widely applied in fields ranging from perturbed eigenvalue problems in quantum mechanics to celestial mechanics \cite{nayfeh1973perturbation, kutz2020advanced}. For oscillatory equations, the standard perturbation method fails to account for the frequency shift introduced by the nonlinearity. To address this limitation, the Lindstedt--Poincar\'e method incorporates frequency corrections directly into the perturbation expansion, systematically eliminating non-physical terms. This approach has since become a foundational tool in celestial mechanics, notably applied to the analysis of the circular restricted three-body problem \cite{ghotekar2019halo, richardson1980analytic}. Beyond astronomy, it has proven valuable in engineering, for example, in vibration analyses of laminated plates \cite{avey2024application} and in biology, where it has been applied to predator–prey models \cite{amore2018application}.

Within PTL-PINNs, such alternative perturbative formulations play a key role in enabling stable and accurate one-shot transfer learning for oscillatory systems and are incorporated naturally into the same computational framework. Using this framework, we demonstrate accurate and efficient solutions across ordinary, coupled and partial weakly nonlinear differential equations. The resulting solutions are expressed as linear combinations of a shared latent representation extracted from a pretrained Multi-Headed-PINN, which directly informs the choice of pretraining equations through the structure of the underlying perturbative system. 

The primary contributions of this work are summarized as follows:

\begin{itemize}
    \item We introduce a Perturbation-Guided Transfer Learning framework for PINNs (PTL-PINNs), which combines perturbation theory with one-shot transfer learning to efficiently solve the sequence of linear systems that approximate weakly nonlinear ODEs and PDEs.
    
    \item We benchmark PTL-PINNs across a broad class of weakly nonlinear systems, including second-order and coupled ODEs, as well as PDEs that are first-order in time and first-order or second-order in space.
    
    \item We show that PTL-PINNs can solve weakly nonlinear systems up to an order of magnitude faster than Runge--Kutta methods by generalizing to new equations via closed-form updates of the output weights, avoiding gradient-based optimization.
    
    \item We derive general $n$th-order perturbation equations and forcing functions using both standard and Lindstedt--Poincar\'e methods. These derivations enable PTL-PINNs to solve perturbative systems to higher orders and guide the selection of pretraining equations.
    
    \item We demonstrate how perturbation methods can be systematically integrated with PINNs. We show that PTL-PINNs reproduce known perturbation-theory failure modes and that the number of perturbation basis functions scales linearly with runtime and logarithmically with error.
\end{itemize}

Viewed broadly, this paper unifies perturbation theory, including the standard perturbation and Lindstedt--Poincar\'e methods, with PINNs, demonstrating how classical analytical structure can be leveraged by modern neural solvers to scale across nonlinear ODEs and PDEs.

The rest of the work is organized as follows. First, we provide an overview of related work in Section \ref{sec:related_work}. Then, we detail the PTL-PINNs methodology in Section \ref{sec:methodology}, describing perturbation theory and one-shot transfer learning. Next, we show its accuracy across a variety of differential equations. Then, we benchmark the computational efficiency of PTL-PINNs in Section \ref{sec:computational_efficiency}. Finally, we discuss the limitations of PTL-PINNs and present future directions in Sections \ref{sec:perturbation_limitations} and \ref{sec:conclusion}.

\section{Related Work}
\label{sec:related_work}

One-shot transfer learning has been proposed to accelerate PINNs for linear equations \cite{desai2024oneshot}, but it is not directly applicable to nonlinear problems. Initial proof-of-concept studies have shown that perturbation theory can be used to extend one-shot transfer learning for weakly nonlinear ODEs \cite{lei2023oneshottransferlearningnonlinear} and PDEs \cite{auroy2025oneshottransferlearningnonlinear} by approximating them with systems of linear equations. The idea of linearizing a problem to accelerate PINNs has also been explored through extreme learning machines, which randomly initialize the hidden-layer weights and only learn the output layer, enabling training via a fast linear optimization process \cite{dwivedi2020physics, VanBeek2025}. 

Other branches of perturbation theory have been leveraged to develop enhanced PINN frameworks. A prominent example is the reformulation as singular perturbation problems of thin boundary layer problems, which are known to challenge both PINNs and traditional numerical methods. Boundary Layer PINNs (BL-PINNs) explicitly incorporate this structure to better resolve boundary layer behavior \cite{arzani2023theory}, with subsequent works demonstrating improved performance and alternative formulations \cite{wang2024general, wang2024chien}. The perturbative Method of Multiple Scales has also been employed to identify appropriate scaling parameters for PINN inputs, leading to more robust multiscale learning \cite{ohashi2024multiple}.

Despite promising approaches that combine perturbation theory with PINNs, the existing literature on perturbation methods primarily focuses on low-order analytical solutions \cite{nayfeh1973perturbation, richardson1980analytic, avey2024application} or relies on semi-symbolic approaches \cite{amore2018application}. Consequently, there is a lack of generalizable frameworks that can systematically integrate PINNs into perturbative analyses. This gap motivates the development of the PTL-PINN framework, which is designed to be broadly applicable to one-shot transfer learning with PINNs. 

\section{Methodology}
\label{sec:methodology}

\subsection{Notation and conventions}

We summarize the notation and indexing conventions used throughout the paper to improve clarity and consistency across perturbative and learning components.

We use $u(t)$ for ODEs and $u(x,t)$ for PDEs. When needed, bold symbols denote vector-valued states. We use $\hat{D}$ and $\mathcal{N}$ to denote abstract linear and nonlinear operators in the general perturbation formulation.
For ODEs, the resulting linear subproblems are written in first-order matrix form using operators $A$ and $B$, with forcing denoted by $F(t)$. For PDEs, the forcing term is written as $f(x,t)$.
When the system is vector-valued, bold symbols (e.g. $\mathbf{u}$, $\mathbf{F}$) are used.
 
Indices and dimensions follow standard conventions.
The perturbation order is indexed by $n=0,\dots,p$.
Training equations (heads) are indexed by $i=1,\dots,k$.
Collocation points are indexed by $i$ (interior), $j$ (initial) and $h$ (boundary).
The latent representation $H$ has dimension $\mathbb{R}^{n \times m}$, where $N$ is the number of collocation points and $m$ the latent width.
Head weights $W_\theta \in \mathbb{R}^{m \times r}$ map the latent space to $r$ state components (or derivatives).
All summations over $t \in T$ denote discrete sums over collocation points.

Throughout this work, $t$ denotes physical time.
In the Lindstedt--Poincar\'e method, we introduce the rescaled time $\tau = \omega t$ to absorb nonlinear frequency shifts.
Here, $\omega_0$ denotes the linear natural frequency, $\omega$ the nonlinear corrected frequency and $\omega_n$ its $n$th-order perturbative correction.
For externally forced systems, $\Omega_k$ denotes prescribed forcing frequencies.

\label{sec:notation}

\subsection{Perturbation theory}
\label{sec:perturbation_theory}

\subsubsection{Introduction}

Perturbation theory deals with weakly nonlinear systems, generally described as:
\begin{equation}
\hat{D}[u] + \varepsilon\, \mathcal{N}[u] \;=\;\mathcal{F}(\mathbf{x}, t),
\label{general_nonlinear}
\end{equation}
Here, $\hat{D}$ denotes a linear differential operator in time and/or space, $\mathcal{N}$ a polynomial nonlinear operator and $\mathcal{F}$ represents a known forcing term incorporating external inputs. 

The small parameter $\varepsilon$ quantifies the deviation from the linear regime. In perturbation theory, the solution is expanded in a series around this coefficient and can be truncated at $p$ terms:

\begin{equation}
    u(x, t) \approx u_0(x, t) + \varepsilon u_1(x, t) + \cdots + \varepsilon^p u^p(x, t)
    \label{eq:x_series_expansion}
\end{equation}

Substituting the expansion in equation \ref{eq:x_series_expansion} into equation \ref{general_nonlinear} and collecting terms of equal powers of $\varepsilon$ yields a sequence of linear differential equations represented in the system of equations \ref{eq:general_linear_system}. The solutions to these linear equations yield the successive correction terms defined in equation \ref{eq:x_series_expansion}:

\begin{equation}
\left\{
\begin{aligned}
    &\mathcal{O}(1): && \hat{D}[u_0] = \mathcal{F} \\
    &\mathcal{O}(\varepsilon): && \hat{D}[u_1] = \mathcal{N}[u_0] \\
    & \quad \vdots \\
    &\mathcal{O}(\varepsilon^p): && \hat{D}[u_p] = - \left( \text{terms of $\mathcal{N}$ of order } \varepsilon^p \right)  \\
\end{aligned}
\right.
\label{eq:general_linear_system}
\end{equation}

This perturbative structure transforms the original nonlinear problem into a sequence of linear subproblems that share the same linear operator but differ in their forcing terms. This structural property is the key feature exploited by the proposed PTL-PINN framework.

\subsubsection{Canonical Oscillator}

To motivate the need for alternative perturbation methods, such as the Lindstedt--Poincar\'e method, we briefly consider the canonical nonlinear oscillator with monomial nonlinearity with power $q$:
\begin{equation}
    u'' + 2 \omega_0 \zeta u' + w^2_0 u + \varepsilon u^q = \sum^m_{k = 0} \Gamma_k \cos(\Omega_k t)
    \label{general_nonlinear_oscillator}
\end{equation}

The linear system that results from inserting the perturbation expansion of equation \ref{eq:x_series_expansion} into equation \ref{general_nonlinear_oscillator} is the following, where the forcing is given by the multinomial coefficient sum:

\begin{equation}
\left\{
\begin{aligned}
    &\mathcal{O}(1): && \ddot{u}_0 + 2 \zeta \omega_0 \dot{u}_0 + \omega_0^2 u_0 = \sum^{m}_ {k= 0} \Gamma _k\cos(\Omega_k t) \\
    &\mathcal{O}(\varepsilon): && \ddot{u}_1 + 2 \zeta \omega_0 \dot{u}_1 + \omega_0^2 u_1 = -u_0^q \\
    & \quad \vdots \\
    &\mathcal{O}(\varepsilon^n): && \ddot{u}_n + 2 \zeta \omega_0 \dot{u}_n + \omega_0^2 u_n \;\; = \sum_{\substack{k_0 + k_1 + \cdots + k_p = q \\ \sum_{i=0}^{p} i\,k_i = n-1}}
\frac{q!}{k_0!\,k_1!\cdots k_p!}\;
\prod_{i=0}^{p} u_i^{k_i} \\
    & \quad \vdots
\end{aligned}
\right.
\label{eq:perturbative_system}
\end{equation}

\subsubsection{Limitations of standard perturbation for oscillatory systems}

The solution of the system of equations \ref{eq:perturbative_system} for the undamped case ($\zeta = 0$) has terms of the form $t^n \sin t$. These terms are non-physical, as they imply unbounded growth in time, despite the underlying system being energy-conserving. This phenomenon is called resonance and occurs in undamped oscillatory systems when the forcing term has a non-zero projection onto the null space of the linear operator, that is, when it shares the same functional form as the homogeneous solution. 

This breakdown reflects a violation of the solvability conditions of the linear subproblems and limits the validity of standard perturbation expansions for oscillatory dynamics. This highlights the need for perturbation formulations that explicitly enforce solvability conditions. While damping suppresses true resonance, near-resonant behavior can still arise in weakly damped regimes due to small-denominator effects \cite{giorgilli1998small}. 

\subsubsection{Lindstedt--Poincar\'e perturbation method}
\label{sec:alternative_perturbation_method}

To ensure the solvability conditions are met when solving undamped oscillators, such as the canonical oscillator and the equilibrium-centered Lotka--Volterra system, we adopt the Lindstedt--Poincar\'e method within the PTL-PINN framework. This prevents resonance, removing non-physical terms from the perturbation series, by introducing a rescaled time variable $\tau = \omega t$, where the frequency $\omega$ is itself expanded as a perturbative series:

\begin{equation}
    \omega \approx \omega_0 + \varepsilon \omega_1 + \cdots + \varepsilon^p \omega_p, \quad \omega \in \mathbb{R}
    \label{eq:frequency_expansion}
\end{equation}

This additional expansion adds flexibility to the series by introducing frequency corrections at each order, which are chosen to eliminate resonant forcing terms. After the time rescaling of the nonlinear equation, each perturbation order yields a modified forcing term that depends on lower-order solutions and the unknown frequency correction.

In our approach, the Lindstedt--Poincar\'e  system is solved numerically rather than symbolically. At each perturbation order $n>0$, the frequency correction $\omega_n$ is computed by enforcing a solvability condition that removes the projection of the forcing onto the homogeneous solution of the linear operator. Once $\omega_n$ is determined, the corresponding linear equation for $u_n$ is solved using PTL-PINNs via one-shot transfer learning, exactly as in the standard perturbation setting.

After calculating the frequency and solution corrections, we sum our series to obtain the final solutions $u \approx u_0 \,+\, \cdots + \varepsilon^n u_n $ and $\omega = \omega_0 + \cdots+ \varepsilon^n \omega_n$. Since we performed the calculations with our original time array, our solution is composed of terms with frequency linear in $t$ and not in $\tau$. By plotting our solution for $t / \omega$, effectively, the frequency of each term is multiplied by $\omega$, ensuring our result captures the nonlinear frequency. 

Importantly, despite the introduction of frequency corrections, the  Lindstedt--Poincar\'e formulation preserves the hierarchical structure required by PTL-PINNs: each perturbation order remains a linear problem sharing the same linear operator, differing only in its forcing term. This allows oscillatory systems to be treated within the same perturbation-guided transfer learning pipeline as non-oscillatory systems, without introducing divergent terms.

Detailed derivations of the  Lindstedt--Poincar\'e perturbative systems for the undamped oscillator and the equilibrium-centered Lotka--Volterra equations are provided in the Appendix.

\subsection{Perturbation-guided transfer learning (PTL-PINNs)}

Both standard perturbation expansions and alternative formulations such as the Lindstedt--Poincar\'e method yield, at each order, linear differential equations sharing a common linear operator but differing in their forcing terms.

\subsubsection{Multi-Headed-PINN}

One-shot transfer learning was proposed by Desai et al. \cite{desai2024oneshot} for linear differential equations of the general form $\hat{D} u(x,t) = f(x,t)$, where $\hat{D}$ is a linear differential operator. This method requires a Multi-Headed-PINN to be pretrained so that it approximates equations of similar form with a shared latent representation, $H(x,t)$. A Multi-Headed-PINN consists of a shared feedforward backbone with $k$ output training equations (heads), each corresponding to a parameter set $\theta^{\mathrm{train}}$. 

Unlike traditional PINN methods, which minimize the loss function of a single equation, the Multi-Headed-PINN pretraining minimizes the average loss across all training equations (heads), defined as $\mathcal{L}_{\text{Multi-Head}} = \frac{1}{k}\sum_{i=1}^{k} \mathcal{L}_i$.

Defining interior collocation points as $\{(x_i,t_i)\}_{i=1}^{N_{\mathrm{int}}}$, 
initial condition points as $\{x_j\}_{j=1}^{N_{\mathrm{ic}}}$, 
and boundary points as $\{t_h\}_{h=1}^{N_{\mathrm{bc}}}$, initial condition as $IC(x)$ and boundary condition as $BC(t)$, the loss for a single equation ($\mathcal{L}_k$) parameterized by $\theta$ is given by:

\begin{equation}
\begin{split}
\mathcal{L}_k =
& \, w_{\mathrm{pde}} \sum_{i=1}^{N_{\mathrm{int}}} ( \hat{D} u_{\theta}(x_i,t_i) - f_{\theta}(x_i,t_i))^{2} \\
& + w_{\mathrm{ic}} \sum_{j=1}^{N_{\mathrm{ic}}} ( u_{\theta}(x_j,0) - IC_{\theta}(x_j))^{2} \\
& + w_{\mathrm{bc}} \sum_{h=1}^{N_{\mathrm{bc}}} ( u_{\theta}(x_h,t_h) - BC_{\theta}(t_h) )^{2}
\end{split}
\label{eq:L_k}
\end{equation}

In the case of ODEs of order $r$, we first rewrite them as a simpler first-order system of the form:
\begin{equation}
    A\,\dot{\mathbf{u}}_{\theta}(t) + B\,\mathbf{u}_{\theta}(t)  = F_{\theta}(t),
    \label{eq:ode_linear_equation}
\end{equation}

Here, $\mathbf{u}_{\theta}(t)$ is the vector of derivatives of $u_{\theta}$ from order $0$ to $r-1$ and $\dot{\mathbf{u}}_{\theta}(t)$ is the vector of derivatives from order $1$ to $r$. The matrices $A$ and $B$ encode the linear coupling between these components and $F_{\theta}(t)$ is the forcing term. 

We adopt this formulation because it provides a unified representation for higher-order and coupled ODE systems within the same architecture. As a result, the canonical oscillator and the coupled Lotka--Volterra system have transformed into the same structural form, allowing both to be handled by the same PTL-PINN model.  

In practice, for $r$-dimensional systems, we split the final latent vector of the shared backbone into $r$ sub-vectors and apply the same linear head to each half. An alternative but mathematically equivalent formulation, where the latent vector is kept intact and two separate linear heads are used, is described in \cite{desai2024oneshot}.

\begin{figure}[H]
    \centering
    \includegraphics[width=1.0\linewidth]{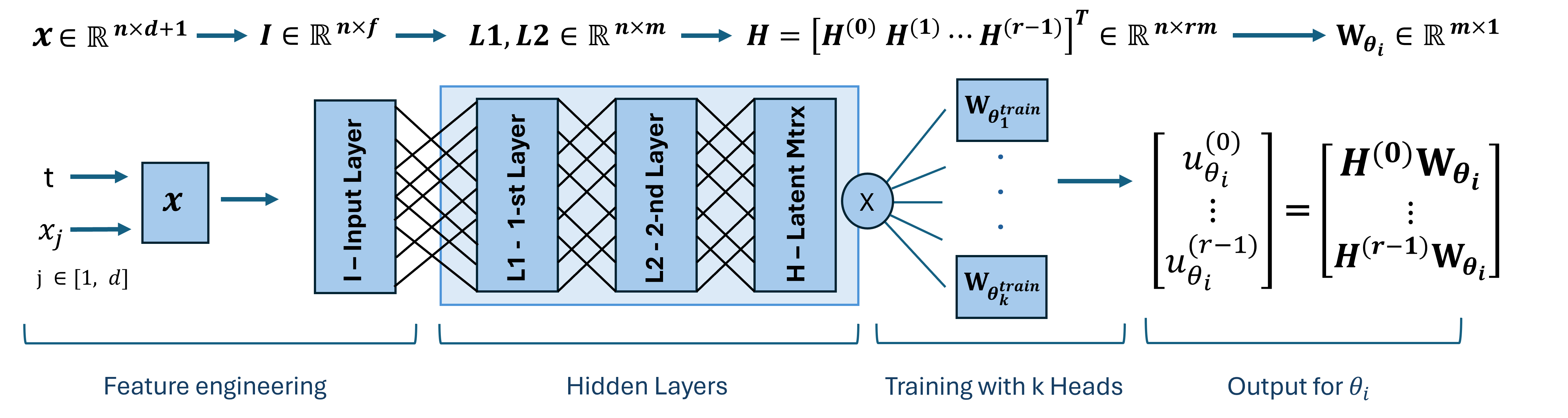}
    \caption{Multi-Headed-PINN architecture used for one-shot transfer learning. A shared neural backbone learns a latent representation $\textbf{H}(x,t)$ across $k$ training equations, while linear output heads represented by matrices $\textbf{W}$ (head weights) map this representation to solution components for each equation. At inference time, new linear problems are solved by computing closed-form output weights without gradient-based optimization.}
    \label{fig:MH-PINN}
\end{figure}

The general architecture we use for pretraining a Multi-Headed-PINN in this manuscript, along with the inference process, is illustrated in Fig. \ref{fig:MH-PINN}. The input $\mathbf{x} \in \mathbb{R}^{n\times d + 1}$ is transformed through a feature engineering stage into $\textbf{I} \in \mathbb{R}^{n \times f}$, which is passed through multiple hidden layers to construct the latent matrix $\textbf{H} \in \mathbb{R}^{n \times rm}$. Finally, through multiplication with $\textbf{H}$, the head weights $W_\theta \in \mathbb{R}^{m \times r}$ map the latent space to $r$ state components (or derivatives).

\subsubsection{One-shot transfer learning}
\label{sec:OSTL}

In this subsection, we outline the derivation of the closed-form expressions used in one-shot transfer learning, following Desai et al. \cite{desai2024oneshot}. After pretraining a Multi-Headed-PINN, the latent representation $H(t, x)$ is extracted and the training-specific head weights $W_{\theta_i^{\mathrm{train}}}$ are discarded. As shown in Fig. \ref{fig:MH-PINN}, the solution for each order parameterized by $\theta$ can be obtained by multiplying our parameter-independent $H(t, x)$ by a head layer $W_{\theta}$. For a new linear differential equation, we parameterize the solution as $\textbf{u}(x,t) = H(x,t) W_{\theta}$ and substitute this form into the loss defined in equation~\ref{eq:L_k}.

\begin{equation}
\begin{split}
\mathcal{L} =
& \, w_{\mathrm{pde}} \sum_{i=1}^{N_{\mathrm{int}}} ( \hat{D} H(x_i,t_i) W_{\theta} - f_{\theta}(x
_i,t_i))^{2} \\
& + w_{\mathrm{ic}} \sum_{j=1}^{N_{\mathrm{ic}}} ( H(x_j,0)W_{\theta} - IC_{\theta}(x_j))^{2} \\
& + w_{\mathrm{bc}} \sum_{h=1}^{N_{\mathrm{bc}}} ( H(x_h,t_h) W_{\theta} - BC_{\theta}(t_h) )^{2}
\end{split}
\label{eq:loss_one_head}
\end{equation}

Since the loss is quadratic in $W_{\theta}$, the optimal head weights are obtained by solving the normal equations corresponding to the condition $\partial \mathcal{L} / \partial W_{\theta} = 0$. Following the derivation of one-shot transfer learning in \cite{desai2024oneshot}, it is possible to compute $W_{\theta}$ for this new equation with equation \ref{eq:calculation_W}.

\begin{equation}
\begin{aligned}
W_{\theta} = M^{-1} \Bigg(&
\sum^{N_{\text{int}}}_{i = 1}
\big(\hat{D}\,H(x_i,t_i)\big)^T f(x_i,t_i) \\
&+ \sum^{N_{\text{ic}}}_{j = 1}
H^T(x_j, 0)\, \text{IC}(x_j) \\
&+ \sum^{N_{\text{bc}}}_{h = 1}
H^T(x_h, t_h)\, \text{BC}(t_h)
\Bigg)
\end{aligned}
\label{eq:calculation_W}
\end{equation}


Here, $M \in \mathbb{R}^{m \times m}$ is the normal matrix arising from the quadratic loss in equation~\ref{eq:L_k} and depends only on the latent representation $H$, the collocation set and on the differential operator $\hat{D}$ .
Its explicit form is provided in \cite{desai2024oneshot}.
Since $M$ is independent of the forcing $F$ and the perturbation order, $M^{-1}$ can be computed once and reused across all perturbation solves. From equation \ref{eq:calculation_W}, the solution can be obtained with time complexity of a matrix inversion or as fast as matrix--vector multiplications if $M^{-1}$ is reused, which can be done within the same perturbation system or between equations with same $\hat{D}$ but different initial/boundary conditions, nonlinearity or forcing.

Obtaining equation \ref{eq:calculation_W} is only possible for linear differential equations. Only these systems lead to a convex optimization problem in the head weights of \(W_{\theta}\), allowing an analytical solution of the $\frac{d\mathcal{L}}{dW_{\theta}} = 0$. Consequently, one-shot transfer learning is limited to linear differential equations and PTL-PINNs are needed to generalize it to weakly nonlinear differential equations.

\subsubsection{Two stages of PTL-PINNs}

\label{sec:methodology_framework}

The PTL-PINN framework consists of two main stages: (1) pretraining a Multi-Headed-PINN on a set of linear differential equations and (2) using the learned latent representation to iteratively solve each linear equation in the perturbative system via one-shot transfer learning.

\paragraph{Stage 1: Perturbation-guided pretraining of a Multi-Headed-PINN}

The PTL-PINN solution is a weighted combination of the latent representation \(H(x,t)\). This motivates the incorporation of the perturbation structure directly into the training process to ensure a meaningful latent representation is found. First, we analyze the structure of the perturbative system associated with the target nonlinear equation. This analysis guides the selection of a representative set of linear differential equations for pretraining the Multi-Headed-PINN. The analytical solutions of the zeroth-order equation guide the choice of the first subset of pretraining equations, while the structure of the first-order forcing informs the selection of additional equations. Finally, we include linear equations with more complex forcing terms that resemble higher-order perturbative contributions, enabling the latent space to generalize to later perturbation orders. The network architecture is selected to balance expressivity and stability across the chosen family of linear problems. The exact equations chosen in training alongside their justification for each system solved can be found in the Appendix.

\paragraph{Stage 2: Iterative perturbation-guided one-shot transfer learning} The solution at perturbation order \( n \) is obtained by solving a linear differential equation parameterized by \( \theta_n \). Each solution is computed by multiplying the parameter-independent latent representation $H(x,t)$ by an output head $W_{\theta_n}$ obtained via one-shot transfer learning. The final solution is constructed by summing contributions from all perturbation orders,  which corresponds to a weighted combination of outputs from the shared latent space \( H(t, x) \):

\begin{equation}
H(t) \;\left( W_{\theta_0} + \varepsilon W_{\theta_1} + \cdots + \varepsilon^p W_{\theta_p} \right)
\label{eq:combination_outputs}
\end{equation}

Provided the initial conditions $x(0)=A$ and $x'(0)=B$, they may be enforced entirely at the zeroth order by setting $x_n(0)=0$ and $x'_n(0)=0$ for $n>0$ in the \textit{leading-order initial condition} approach. Alternatively, the initial conditions may be distributed uniformly across perturbation orders in the \textit{uniform initial condition} approach by setting
\[
x_n(0) = \frac{A}{\sum_{k=0}^p \varepsilon^k}, \quad
x'_n(0) = \frac{B}{\sum_{k=0}^p \varepsilon^k}.
\]

\subsubsection{Initial condition handling in PTL-PINNs}

\begin{figure}[H]
	\centering
\includegraphics[width=1.0\linewidth]{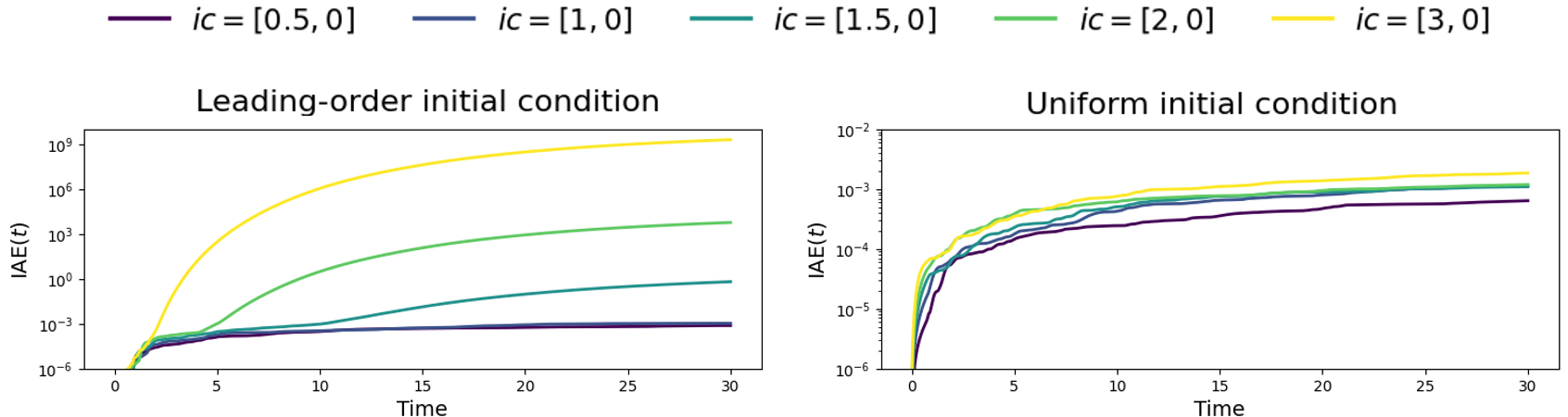}

\caption{Comparison of initial-condition handling strategies for the unforced overdamped oscillator ($\zeta = 10$).
PTL-PINN solutions are shown using 15 perturbation corrections for a cubic nonlinearity $0.5x^3$.
The \emph{leading-order} approach enforces the full initial condition at zeroth-order only, while the \emph{uniform} approach distributes the initial condition evenly across all perturbation orders.
Results are shown for increasing initial positions $x(0) \in \{0.5, 1, 1.5, 2, 3\}$ with zero initial velocity.
}
\label{fig:overdamped_different_ics}
\end{figure}

In Fig. \ref{fig:overdamped_different_ics}, we compare the  \textit{leading-order initial condition}  and \textit{uniform initial condition} by solving with PTL-PINNs the unforced overdamped oscillator for various initial conditions using fifteen correction terms for a nonlinearity of $0.5 x^3$ and damping $\zeta = 10$. 

An overdamped regime was chosen because, unlike the undamped and underdamped regimes, it does not exhibit resonance or near-resonance behavior, as its homogeneous solution is purely exponential. Despite this, the perturbation series will still diverge for large initial conditions. This limitation is inherent to perturbation theory and arises because the perturbation terms are multiplied and raised to high powers throughout the perturbation system. It can be mitigated by adopting alternative approaches to the treatment of initial conditions.

The \textit{uniform} approach leads to low errors for larger initial conditions. By distributing the initial condition evenly across all orders, the starting amplitude of lower-order corrections is kept within a manageable range. This delays the onset of a blow-up, which is largely determined by the amplitude of $x_0$. However, the \textit{leading-order} approach has various advantages. It produces a truncatable series, where the contribution of each order is independent of the initial conditions. In contrast, the \textit{uniform} approach cannot be truncated without losing information, since satisfying the initial condition requires including all corrections. Due to its interpretability, the results shown throughout this manuscript have considered the \textit{leading-order} approach.

\subsection{Importance of PTL-PINN pretraining}

\begin{figure}[H]
    \centering
    \includegraphics[width=1.0\linewidth]{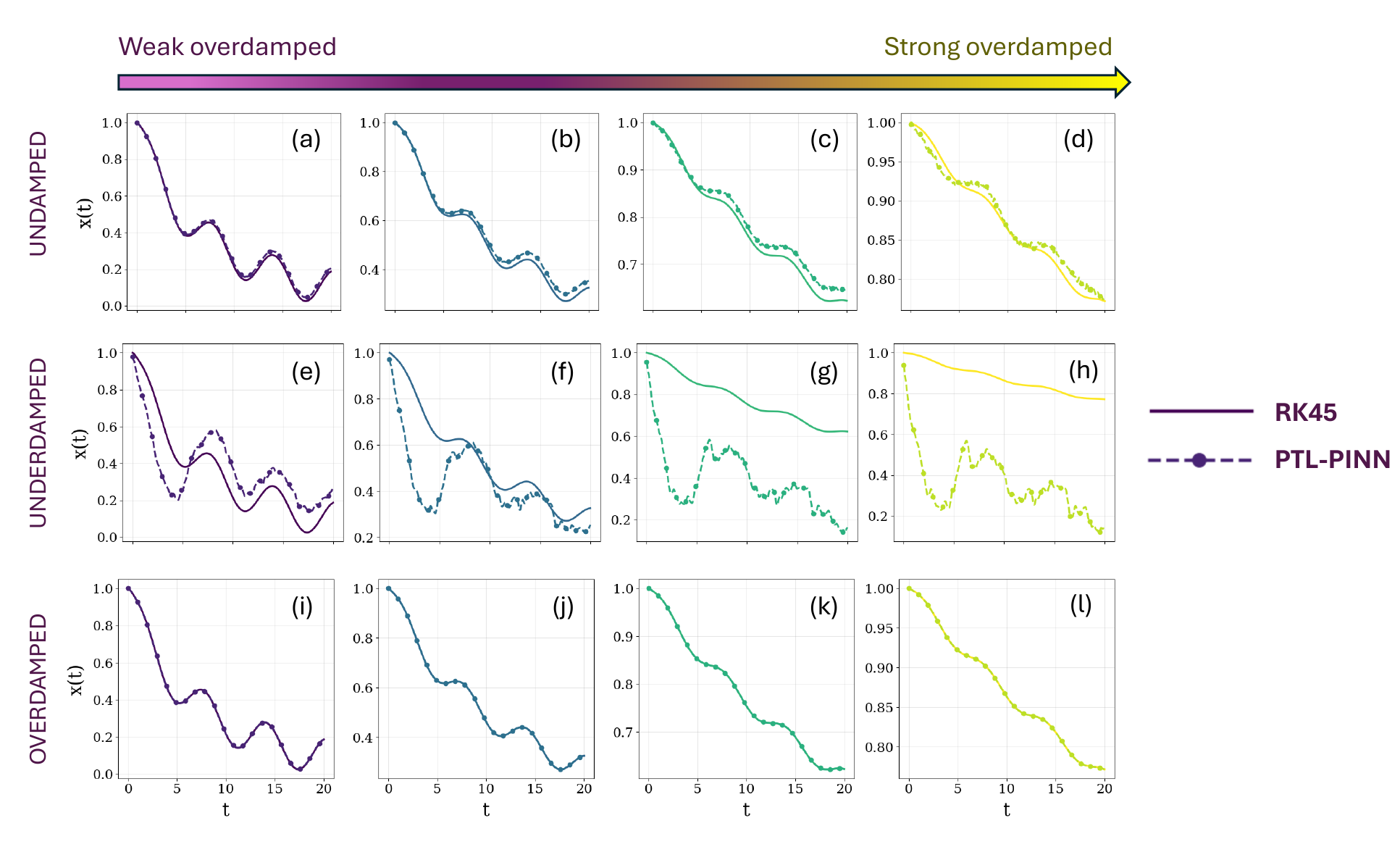}
    \caption{Comparison of the solution of various nonlinear overdamped oscillators using the baseline RK45 and PTL-PINN undamped, underdamped and overdamped models. Each row corresponds to a fixed PTL-PINN model, while each column represents a different damping value from $\{5, 10, 30, 60\}$. The PTL-PINN overdamped model has greatest accuracy.}
    \label{fig:overdamped_error}
\end{figure}

In one-shot transfer learning, we obtain the solution by taking a linear combination of the latent representation as described in equation \ref{eq:combination_outputs}. As such, the perturbation-guided pretraining stage is crucial to ensure a meaningful latent representation is learned. In this section, we demonstrate the importance of the pretraining stage. Fig. \ref{fig:overdamped_error} shows the results of using nine corrections to solve various forced overdamped oscillators with models trained on undamped (a)-(d), underdamped (e)-(h) and overdamped (i)-(l) linear equations.

Fig. \ref{fig:overdamped_error} (i)-(l) shows that the PTL-PINN trained on overdamped equations accurately reconstructs the nonlinear RK45 reference solution. In contrast, the undamped and underdamped models are unable to reconstruct the overdamped solution, which highlights the need to train a model for each regime.

\begin{figure}[H]
	\centering
\includegraphics[width=0.90\linewidth]{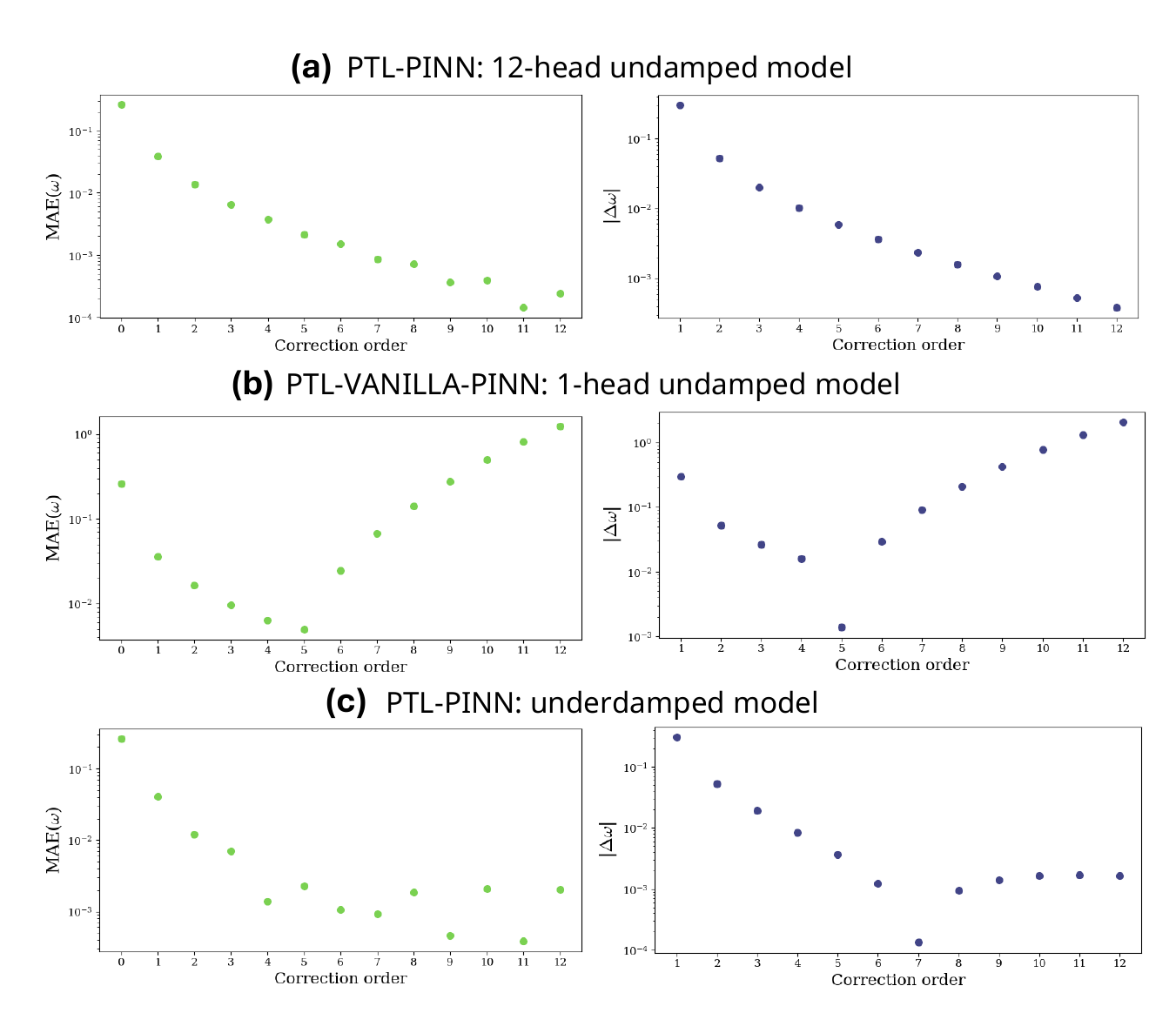}

\caption{Lindstedt--Poincar\'e frequency convergence for the undamped and unforced oscillator with $x(0) = 1$ and $x'(0) = 0$ using an undamped (twelve-headed), underdamped and vanilla (one-headed) model, showing absolute relative frequency contribution (right) and
frequency series MAE (left)}
	\label{fig:LPM_convergence}
\end{figure}

Besides the parameters considered for the pretraining equations, the number of pretraining equations (heads) directly impacts the results. Fig. \ref{fig:LPM_convergence} shows how well different PTL-PINN models can recover Lindstedt--Poincar\'e frequency series for the unforced undamped oscillator, described in equation \ref{eq:frequency_expansion}. The left plots in Fig. \ref{fig:LPM_convergence} show the absolute relative frequency contribution, which quantifies the impact of each frequency correction on the series, as more perturbation corrections are considered. The right plots show the frequency series $\mathrm{MAE}$, where RK45 provides reference frequency, computed from the peak-peak distance over the short interval \( [0, 4\pi] \). To maximize accuracy, we consider with absolute and relative tolerances of $10^{-10}$ for RK45.

Fig. \ref{fig:LPM_convergence} (a) shows the results for a twelve-headed undamped PTL-PINN model, which achieved the lowest $\mathrm{MAE}(\omega)$ and $|\Delta \omega|$, both decreasing steadily with each order. Fig. \ref{fig:LPM_convergence} (b) presents the results for a one-headed undamped PTL-PINN. Trained on a single equation, the model’s latent representation is not expressive enough to capture the underlying dynamics, leading to frequency divergence beyond the fifth correction. Fig. \ref{fig:LPM_convergence} (c) shows the results for a PTL-PINN trained on twelve underdamped systems. As these pretraining equations differ from the target dynamics, the resulting latent representation lacks generalizability, leading to frequency divergence beyond the seventh correction.

We notice there is a correlation between the relative frequency and its absolute error. When the relative frequency does not diminish as we consider more corrections, we find that adding more corrections does not necessarily imply a more accurate perturbation series with lower absolute error. As a result, a steadily diminishing relative frequency provides a stopping criterion for our algorithm. This allows divergence to be identified when solving our perturbation system and early truncation. 

\begin{figure}[H]
    \centering
    \includegraphics[width=1\linewidth]{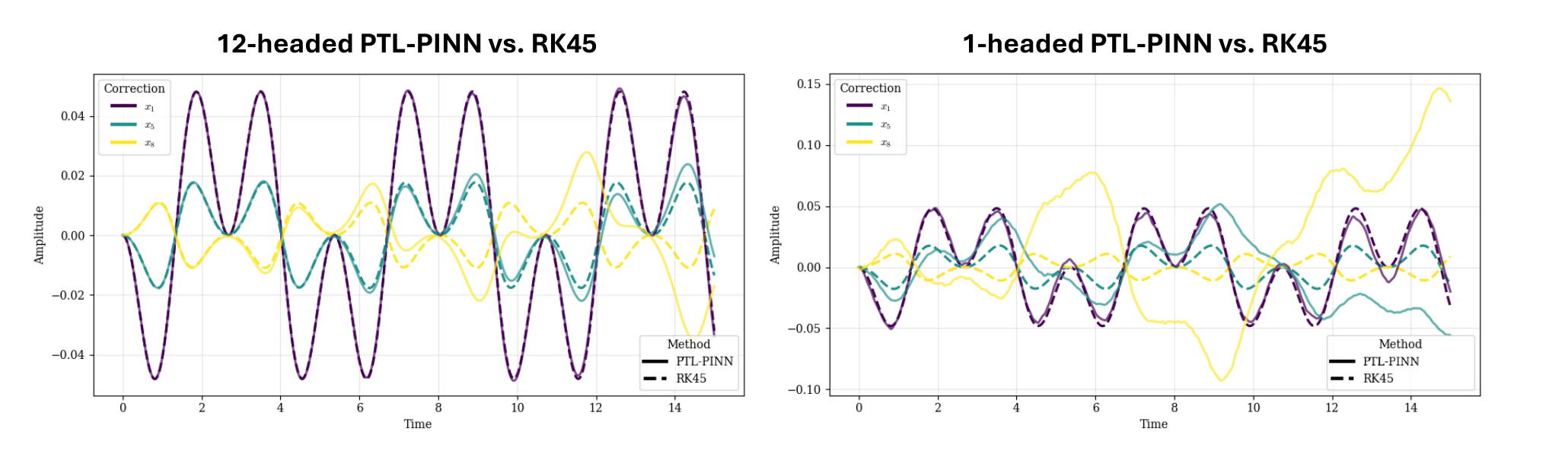}
    \caption{Comparison of the first, fifth and eighth-order corrections when using RK45 and a twelve-headed and one-headed PTL-PINN model.}

\label{fig:perturbation_system_comparison}
\end{figure}

To further benchmark the results of our PTL-PINN undamped model, we solve the Lindstedt--Poincar\'e perturbative system of the unforced undamped oscillator with RK45 for nonlinearity $0.5 x^3$ and initial conditions $x(0) = 1$ $\dot{x}(0) = 0$. Fig. \ref{fig:perturbation_system_comparison} compares the solution of the first, fifth and eighth-order corrections from RK45, a PTL-PINN twelve-headed model and a PTL-PINN one-headed model. 

These results show that the twelve-headed PTL-PINN has excellent agreement with the RK45 solution for the first and fifth-order corrections, with a larger deviation observed for the eighth-order correction. Conversely, a one-headed PTL-PINN model is only able to find the first-order correction without significant total error. Fig. \ref{fig:perturbation_system_comparison} demonstrates that PTL-PINN accuracy is highly dependent on the number of pretraining equations (heads), being limited by the expressiveness of the latent representation. Moreover, these results show the challenge PTL-PINNs face in generalizing to higher-order corrections, especially for models trained on a small number of equations or equations unrelated to the system being solved.

\section{Accuracy Results on Nonlinear Systems}

\subsection{Numerical solver baselines}
\label{sec:numerical_baselines}

We compare the accuracy of PTL-PINNs against the RK45 solution of the nonlinear equation, since it is a widely used explicit method with fourth-order accuracy \cite{butcher1996history}. To further benchmark accuracy, we also solve the Lindstedt--Poincar\'e and standard perturbation systems using RK45 solely to assess how accurately the PTL-PINN reproduces the perturbation-theory solutions across various orders of perturbation. We evaluate accuracy using the mean absolute error ($\mathrm{MAE}$) and the integrated absolute error ($\mathrm{IAE}$):

\begin{equation}
    \mathrm{MAE}(x_{\text{approx}}) = \frac{1}{N} \sum_{i=0}^{N-1} \left| x_{\text{approx}}(i) - x_{\text{reference}}(i) \right|
\end{equation}
\begin{equation}
    \mathrm{IAE}(t) = \int_0^t \left| x_{\text{approx}}(m) - x_{\text{reference}}(m) \right| \, dm
\end{equation}

To assess computational efficiency, we also consider Radau, an implicit Runge-Kutta method particularly well-suited for stiff systems \cite{hairer1991ii}. For PDEs, we first discretize the spatial domain and then apply the time integration method RK45, which operates on the resulting system of ordinary differential equations obtained from the finite-difference discretization.

The numerical solvers considered, RK45 and Radau, have adaptive time-stepping. For a given relative and absolute tolerance, these solvers adjust the time step to ensure the local error estimate remains less than the sum of the absolute tolerance plus the relative tolerance multiplied by the absolute value of the solution. As such, for any accuracy baselines, we ensure high accuracy by setting absolute and relative tolerances of $10^{-10}$. For our computational efficiency benchmarks, we consider tolerances of $10^{-3}$ for RK45 and Radau, which we demonstrate to be comparable to the accuracy achieved by PTL-PINNs.

\subsection{ODEs: Damped and Undamped Oscillator}

\label{sec:undamped_standard}

The expected structure of the perturbative solution for the nonlinear oscillator depends on the damping regime. To ensure that a meaningful latent representation is learned, we train a separate PTL-PINN model for each of the undamped, underdamped and overdamped cases, as described in Section~\ref{sec:training_model}. 

PINNs often struggle with spectral bias in training. To mitigate this, when solving the undamped and underdamped oscillator, we use Fourier feature embeddings \cite{wang2021eigenvector}, which map the inputs through sinusoidal functions with various uniformly sampled frequencies. Furthermore, we use sinusoidal activation functions for the first two layers of the undamped and underdamped models \cite{sitzmann2020implicit} and for the first layer of the overdamped model.

\begin{figure}[H]
    \centering
    \includegraphics[width=1\linewidth]{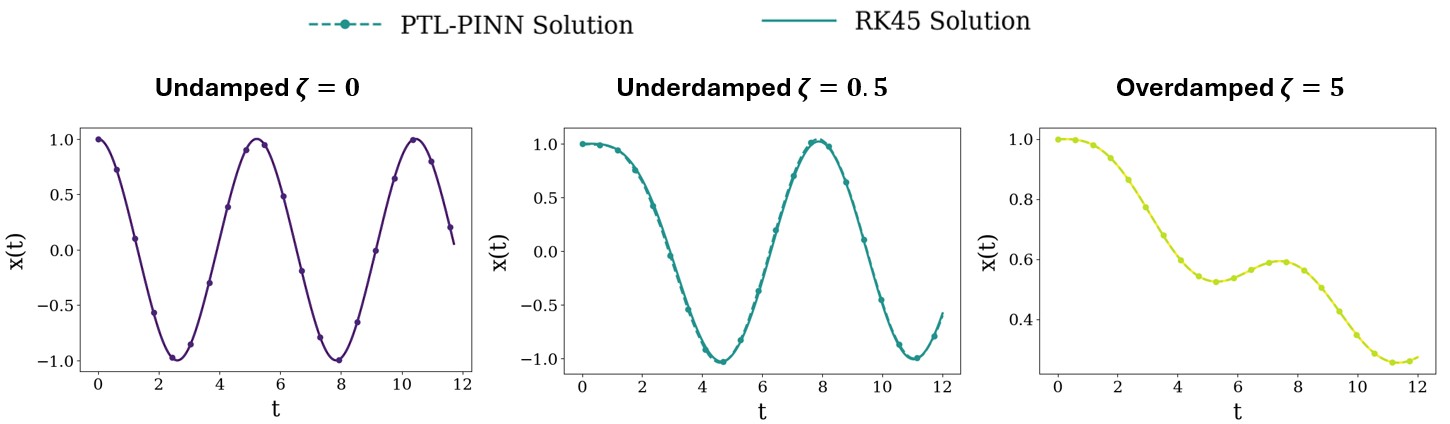}
    \caption{PTL-PINN vs. baseline RK45 solutions for the unforced and undamped oscillator and the forced underdamped ($\zeta = 0.5$) and overdamped ($\zeta = 5$) oscillators under external forcing $\cos(t)$. PTL-PINN solutions is shown to be accurate using Lindstedt--Poincar\'e method for the undamped oscillator and the standard perturbation method for the damped oscillators.}

    \label{canonical_oscillator}
\end{figure}

In Fig. \ref{canonical_oscillator}, we show that PTL-PINNs using five corrections are able to recover, with $\mathrm{MAE}$ of order $10^{-2}$ or less, the unforced undamped oscillator using the Lindstedt--Poincar\'e method and the underdamped ($\zeta = 0.5$) and overdamped ($\zeta = 5$) oscillators forced by $\cos(t)$  using the standard perturbation method for the damped cases. 

We consider initial conditions $x(0) = 1$, $\dot{x} = 0$ and a polynomial nonlinearity of $-x^3 + x^5$ with the coefficient $\varepsilon = 0.5$, which corresponds to a quintic approximation of the nonlinearity $\sin(x)$. PTL-PINNs naturally extend to nonlinearities of higher polynomial order within the same framework.

\subsection{Coupled ODEs: Predator–prey Models}

\label{sec:intro_lotka_volterra}

The Lotka--Volterra system describes the nonlinear interaction between two species in a predator--prey system. Their normalized version is given by:
\begin{equation}
\begin{aligned}
    \dot{x} &= x - x y,\\
    \dot{y} &= - \alpha y + \alpha x y,
\end{aligned}
\label{eq:LV}
\end{equation}
where $x(t)$ and $y(t)$ represent the prey and predator populations and $\alpha > 0$. This system admits a non-trivial steady state $(x^*,y^*) = \left( 1, 1\right)$. We apply PTL-PINNs in combination with the Lindstedt--Poincar\'e method to approximate solutions of the Lotka--Volterra system in a neighborhood of the non-trivial steady state. 

Following Grozdanovski weakly nonlinear formulation of the Lotka--Volterra system \cite{grozdanovski2007approximating}, we introduce perturbative variables $\xi$ and $\eta$ defined by $x(t) = 1 \,+ \, \varepsilon \xi(t)$ and $y(t) = 1 \, + \, \eta(t)$. Then, we define their perturbative series as $\xi = \xi_0 + \xi_1 \varepsilon + \cdots$ and $\eta = \eta_0 + \eta_1 \varepsilon + \cdots$ and implement the Lindstedt--Poincar\'e method numerically similarly to how we approached the undamped oscillator. More details on the derivation of the general $n$-th order differential equation and frequency correction can be found in the corresponding methodological subsection present in  \ref{sec:LK_LPM}.

\begin{figure}[H]
    \centering
    \includegraphics[width=1.0\linewidth]{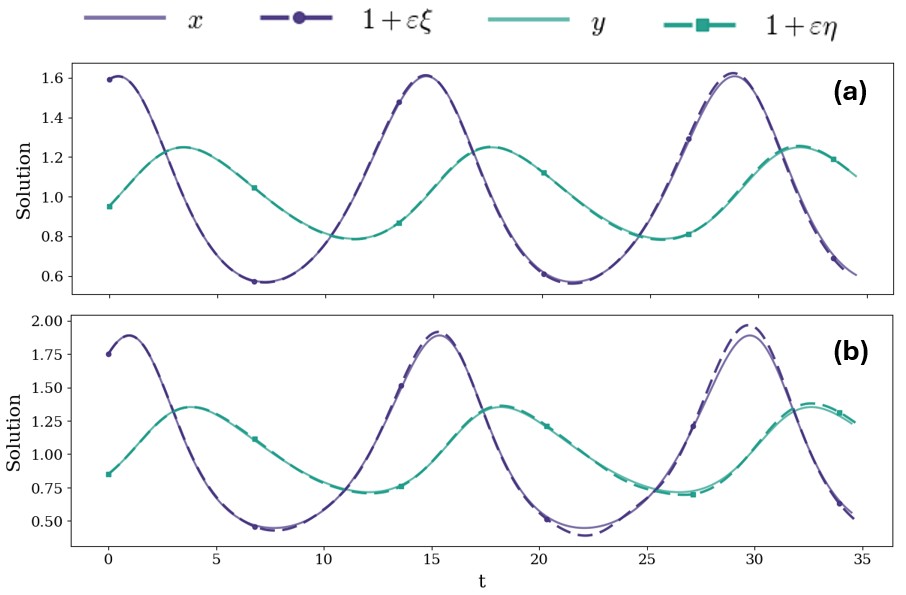}
    \caption{
    PTL-PINN (dashed line) vs. RK45 baseline (continuous line) solutions of the equilibrium-centered Lotka–Volterra system for $\alpha = 0.2$  and  $\epsilon = 0.5$. The same latent representation trained on the undamped oscillator is reused to recover periodic orbits around the non-trivial equilibrium $(1,1)$ using the Lindstedt--Poincar\'e method. 
Panels (a) and (b) correspond to increasing deviations from equilibrium with initial conditions $x(0)=1.59, \,\, y(0)=0.95$ for (a)
and $x(0)=1.75, \,\, y(0)=0.85$ for (b).}
    \label{fig:Lotka_Volterra}
\end{figure}

Because the functional form of the perturbative solutions resembles that of the undamped oscillator, we reuse the PTL-PINN model trained for the undamped oscillator to solve the Lotka--Volterra for $\alpha = 0.2$ and $\varepsilon = 0.5$. We consider initial conditions $x(0) = 1.59$ and $y(0) = 0.95$ in Fig. \ref{fig:Lotka_Volterra} (a) and $x(0) = 1.75$ and $y(0) = 0.85$ in Fig. \ref{fig:Lotka_Volterra}  (b). 

Fig. \ref{fig:Lotka_Volterra} shows that PTL-PINNs combined with the Lindstedt--Poincar\'e method can accurately recover the periodic orbits around the non-trivial critical point $(x^*, y^*) =(1,1)$. 

Because the Lindstedt--Poincar\'e expansion is constructed around this equilibrium, we expect its accuracy to deteriorate as the initial condition moves farther from this point. In fact, PTL-PINN achieves $\mathrm{MAE}$ of $8 \times 10^{-3}$ for $\xi(0) = 1.59$ and $\eta(0) = 0.95$. However, for $\xi(0) = 1.75$ and $\eta(0) = 0.85$, it achieves a lower $\mathrm{MAE}$ of $3 \times 10^{-2}$ as this point further is from (1,1). PTL-PINNs can also be applied in regimes of weak coupling between species, where the cross-species interaction term ($xy$) remains small relative to each population’s self-dynamics.

\subsection{PDEs: KPP--Fisher and Wave Equation}

Reaction-diffusion and wave phenomena appear across a wide range of systems and can be modeled by the KPP--Fisher and Wave equations. The KPP--Fisher equation models population growth and wave propagation \cite{zou2002delay, brunet2016some}. To apply the PTL-PINN framework, we reformulate the systems in their weakly nonlinear form in accordance with PTL-PINNs' general formulation presented in equation \ref{general_nonlinear}. 

Following this formulation, the KPP--Fisher weakly nonlinear form is given by equation \ref{eq:weak_kpp}, where the linear operator is $D$ is the diffusion coefficient, $\hat{D} = \frac{\partial}{\partial t} - D \frac{\partial^2}{\partial x^2}$ is the linear differential operator and $\mathcal{N} = - \, u \,(1 -u)$ is the nonlinear operator:
\begin{equation}
    \frac{\partial u}{\partial t}
    - D \frac{\partial^2 u}{\partial x^2}
    - \varepsilon\,u(1 - u)
    = f(x,t)
    \label{eq:weak_kpp}
\end{equation}

\begin{figure}[H]
    \centering
    \includegraphics[width=1.0\linewidth]{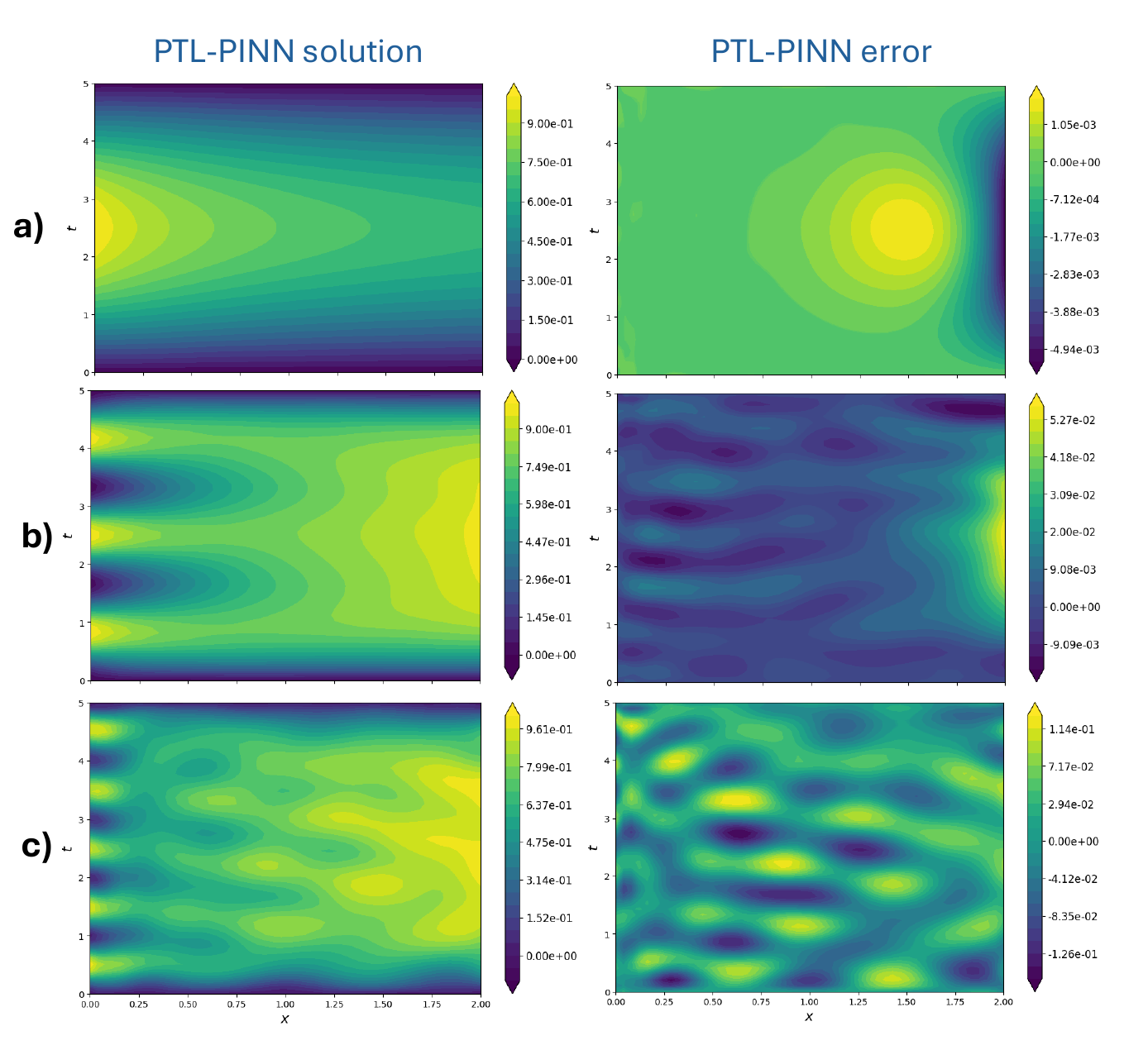}
    \caption{PTL-PINN solutions (left panel) and errors (right panels) for the weakly nonlinear KPP--Fisher equation for different initial conditions. The initial conditions are progressively more complex from (a)-(c), taking higher frequency and/or more harmonics. Consequently, the error increases from (a)-(c) as it becomes harder for PTL-PINN to generalize. $\mathrm{MAE}$ values are $4.6 \times 10^{-4}, 6.3 \times 10^{-3}$  and $4.0 \times 10^{-2}$ for (a), (b) and (c), respectively.}
    \label{eq:PDE_KPP}
\end{figure}

Similarly, the Wave equation weakly nonlinear formulation is given by equation \ref{eq:weak_wave}, where the linear wave speed is $c$, the linear operator is $\hat{D} = \frac{\partial^2}{\partial t^2} - c^2 \frac{\partial^2}{\partial x^2}$ and the nonlinear operator is $\mathcal{N} = u^q$.

 and its weakly nonlinear formulation is:
\begin{equation}
    \frac{\partial^2 u}{\partial t^2}
    - c^2 \frac{\partial^2 u}{\partial x^2}
    + \varepsilon\, u^q
    = f(x,t),
    \label{eq:weak_wave}
\end{equation}
We solve these equations for $x \in (0, 2)$ , $t \in (0, 5)$ and $\varepsilon = 0.5$. The training equations are provided in \ref{sec:training_config_pdes} and are guided by perturbation theory as discussed earlier. Fig. \ref{eq:PDE_KPP} shows PTL-PINN solutions of the KPP--Fisher for Dirichlet boundary and initial conditions of varying complexity and frequency. In Fig. \ref{eq:PDE_KPP} (a), we solve for the initial condition $\sin(2\pi x/L)$ with diffusion coefficient $D = 0.1$. Fig. \ref{eq:PDE_KPP} (b), the solution was obtained for $D = 0.01$ and the initial condition $\sin^{2}(3\pi x/L)$, while Fig. \ref{eq:PDE_KPP}(c) was computed using $D = 0.01$ with the initial condition $\sin^{2}(5\pi x/L)$. As the complexity of the initial condition increases from (a) to (c), the corresponding $\mathrm{MAE}$ values increase, being $4.6 \times 10^{-4}$, $6.3 \times 10^{-3}$ and $4.0 \times 10^{-2}$ for (a), (b) and (c), respectively.

\begin{figure}[H]
    \centering
    \includegraphics[width=1.0\linewidth]{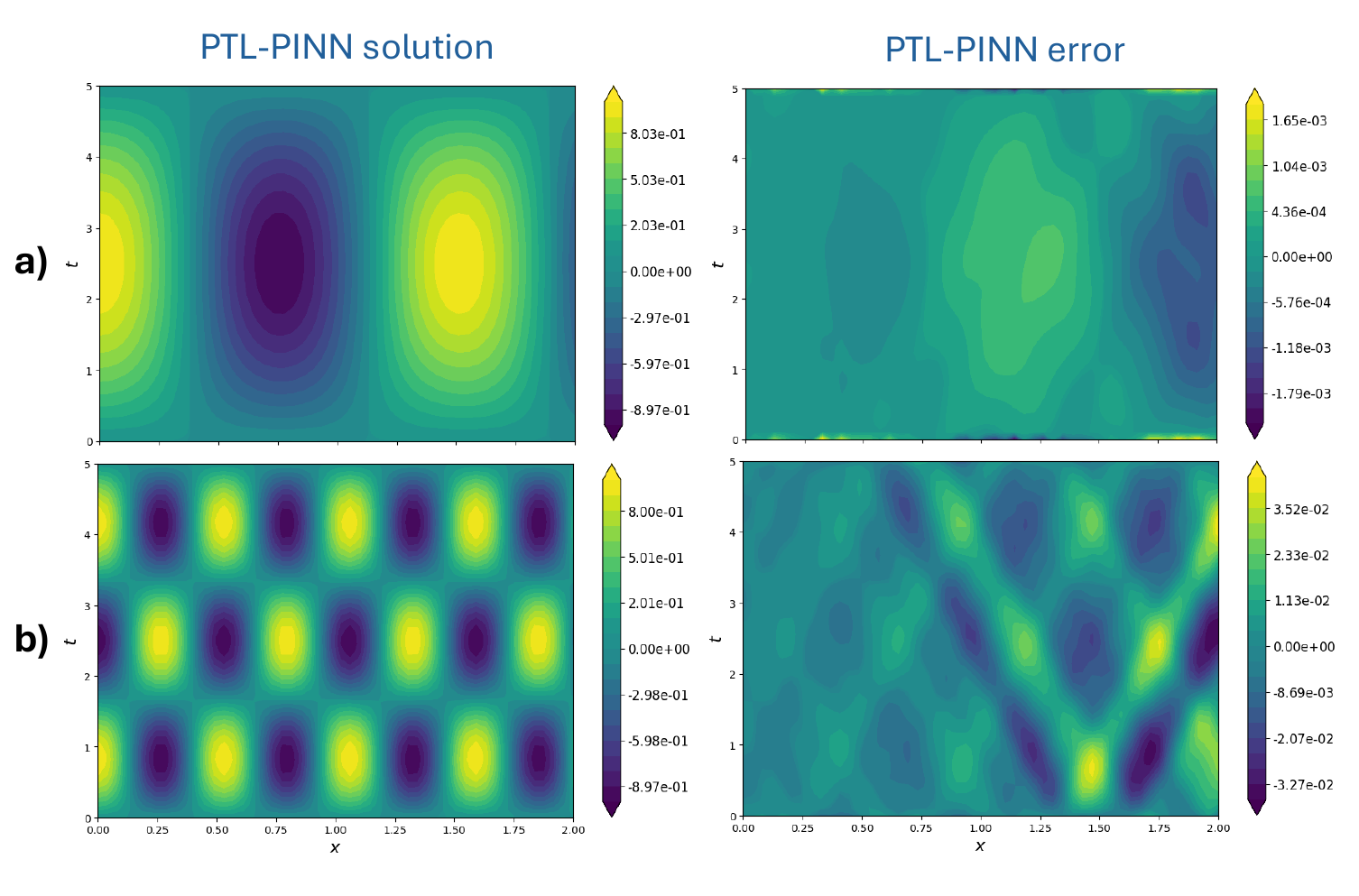}
    \caption{PTL-PINN solutions (left panel) and errors (right panels) for the weakly nonlinear Wave equation with Dirichlet
boundaries, linear wave velocity $c = 1$ and  
equation for different initial conditions. Panel (a) corresponds to the initial condition $\sin(2\pi x/L)$ and panel (b) corresponds to the Wave equation with an initial condition $\sin(3\pi x/L)$. As the frequency of the initial condition increases from (a)
to (b), the corresponding MAE values increase, indicating higher
error for the higher frequency condition.}   \label{eq:PDE_Wave}
\end{figure}

The same is replicated for the nonlinear Wave equation with nonlinearity of power $q = 3$. Fig. \ref{eq:PDE_Wave} shows the PTL-PINN solution for the Wave equation, considering Dirichlet boundaries, linear wave velocity of $c = 1$ and initial conditions of varying frequency. In Fig. \ref{eq:PDE_Wave}(a), we consider the initial condition $\sin(2\pi x/L)$. Fig. \ref{eq:PDE_Wave}(b) corresponds to the initial condition $\sin(3\pi x/L)$. As the frequency of the initial condition increases from (a) to (b), the corresponding $\mathrm{MAE}$ values increase, being $2.9 \times 10^{-4}$ and $7.9 \times 10^{-3}$ for (a) and (b), respectively.

In this subsection, we demonstrated the success of PTL-PINNs on benchmark PDEs and their ability to scale to more complex initial conditions. This framework can be naturally scaled to more complex PDEs and higher frequencies by adjusting the model and training equations. This can be extended to any PDE, first requiring its weakly nonlinear formulation. Various PDEs, such as the Navier-Stokes, have been solved with perturbation theory \cite{pierson1962perturbation}.

\section{Computational Efficiency}
\label{sec:computational_efficiency}

PTL-PINNs solve each linear equation in the perturbative system by multiplying a precomputed latent representation $H(t, x)$ by a matrix $W_{\theta}$, which is obtained via the closed-form expression in equation \ref{eq:calculation_W}. The computation of $W_{\theta}$ involves matrix multiplications and the inverted matrix $M^{-1}$. As discussed in Section \ref{sec:OSTL}, $M^{-1}$ is independent of the initial/boundary conditions, forcing function or nonlinearity. Consequently, the same matrix $M^{-1}$ can be reused across all perturbation orders in the same system, since only the initial/boundary conditions and forcing terms vary. Moreover, it can be reused between nonlinear equations, where only the forcing, initial conditions or nonlinearity varies. If $M^{-1}$ is unavailable, the solution is found with a time complexity equivalent to that of a matrix inversion. If it is, then the solution is even faster with the time complexity of matrix--vector multiplications.

Table \ref{tab:ptl_time_vs_classical_zeta} compares the solve time of the unforced case of the PDEs and ODEs previously discussed using: PTL-PINNs (reusing and inverting $M$), classical solvers (RK45 and Radau) and regular transfer learning (Regular-TL). In Regular-TL, we trained a Multi-Headed-PINN on various versions of the nonlinear oscillator that we solve. We freeze the shared backbone of the network and find the final layer $W$ with gradient-based optimization. The computational time expresses how long it took for Regular-TL to reach $\mathrm{MAE} < 2.5 \times 10^{-2}$, when compared to a low absolute tolerance RK45 solution. Our results show that Regular-TL is unreliable, taking many orders of magnitude longer than other solvers, only to find a less accurate solution. Conversely, the PTL-PINN solutions can be up to one order of magnitude faster than classical solvers.

\newcolumntype{P}[1]{>{\raggedright\arraybackslash}p{#1}}

\begin{table}[H]
\centering
\small
\setlength{\tabcolsep}{8pt}
\renewcommand{\arraystretch}{1.2}

\resizebox{\textwidth}{!}{%
\begin{tabular}{P{5.0cm}
                p{2.0cm}   
                p{2.0cm}   
                p{2.5cm} 
                p{3cm}   
                p{3.6cm}}
\toprule
& \multicolumn{5}{c}{\textbf{Time (milliseconds)}} \\
\cmidrule(lr){2-6}
\textbf{Equation} &
\textbf{RK45} &
\textbf{Radau} &
\textbf{Regular TL*} &
\textbf{PTL-PINN-invert} &
\textbf{PTL-PINN-no-invert} \\
\midrule
Undamped             & 28.9 & 11.1 & 2.4 \(\times 10^{4}\) & 70.2 & 4.8 \\
Underdamped (\(\zeta=0.4\)) & 4.7 & 1.2  &  1.4 \(\times 10^{5}\) & 19.1 & 1.1 \\
Underdamped (\(\zeta=0.6\)) & 3.7 & 1.4  &  1.8 \(\times 10^{3}\)  & 14.0 & 1.1 \\
Overdamped (\(\zeta=10\))     & 3.6  & 6.4 & 3.23 \(\times 10^{4}\) & 21.7 & 1.1 \\
Overdamped (\(\zeta=30\))     & 2.0  & 22.8 & 7.75 \(\times 10^{5}\) & 14.1 & 1.1 \\

Fisher–KPP        & 47.7 & 6.2 & $6.5 \times 10^{3}$                     & 311.1  & 15.0     \\
Wave equation     & 64.4     & 263.8     & $8.3 \times 10^{3}$                     &  396.5  & 18.3    \\
Lotka–Volterra ($\alpha = 0.2$) & 1.3 &  6.0  & 3.15 \(\times 10^5\) & 62.9 & 6.3 \\
\bottomrule
\end{tabular}%
}
\caption{Solve time for PTL-PINNs (inverting and not inverting $M$), Regular TL, Radau and RK45 for various equations. The asterisk (*) denotes that Regular TL times were obtained when the solution matches the RK45 numerical solution with a tolerance of $2.5 \times 10^{-2}$, which is a substantially larger error than that obtained with the other solvers.}

\label{tab:ptl_time_vs_classical_zeta}
\end{table}

The equations were solved for $t \in (0, 10)$, $\varepsilon = 0.5$ and without forcing. We use a monotonic clock with the highest available resolution to measure the time and average eight hundred iterations of our numerical solvers and PTL-PINN results. For PTL-PINNs, we exclude the one-time cost of computing $H(t)$ and derivatives since they can be reused. As we will later justify in this section, we consider five correction terms, $N = 150$ inference points for ODEs and $N = 50 \times 50$ points for PDEs. As discussed in Section \ref{sec:numerical_baselines}, we set the tolerances of RK45 and Radau to $10^{-3}$ for fair comparison of computational efficiency. Table \ref{tab:accuracy_comparisons} shows the $\mathrm{MAE}$ achieved for the solvers presented in Table \ref{tab:ptl_time_vs_classical_zeta}, where the reference is RK45 with tolerance of $10^{-10}$. The values of RK45 and Radau consider the same tolerance as in Table \ref{tab:ptl_time_vs_classical_zeta}, which is $10^{-3}$.


\begin{table}[H]
\centering
\small
\setlength{\tabcolsep}{8pt}
\renewcommand{\arraystretch}{1.2}

\resizebox{\textwidth}{!}{%
\begin{tabular}{
    P{5.0cm}   
    p{2.0cm}   
    p{2.0cm}   
    p{2.5cm}   
    p{3.6cm}   
}
\toprule
& \multicolumn{4}{c}{\textbf{Accuracy ($\mathrm{MAE}$)}} \\
\cmidrule(lr){2-5}
\textbf{Equation} &
\textbf{RK45} &
\textbf{Radau} &
\textbf{Regular TL*} &
\textbf{PTL-PINN} \\
\midrule
Undamped      & $1.1 \times 10^{-2}$ & $7.6 \times 10^{-5}$ & $2.5 \times 10^{-2}$ & $2.2 \times 10^{-3}$ \\
Underdamped ($\zeta = 0.4$) & $6.9 \times 10^{-4}$ & $4.1 \times 10^{-5}$ & $2.5 \times 10^{-2}$ & $4.8 \times 10^{-3}$ \\
Underdamped ($\zeta = 0.6$) & $9.1 \times 10^{-4}$ & $3.4 \times 10^{-5}$ & $2.5 \times 10^{-2}$ &  $8.1 \times 10^{-3}$ \\
Overdamped ($\zeta = 10$)   & $2.5 \times 10^{-5}$ & $2.9 \times 10^{-5}$ & $2.5 \times 10^{-2}$ & $1.5 \times 10^{-4}$ \\
Overdamped ($\zeta = 30$)  & $6.0 \times10^{-6}$ & $8.2\times10^{-6}$ & $2.5 \times 10^{-2}$ & $4.2 \times 10^{-4}$ \\
Fisher–KPP    & $4.3 \times 10^{-4}$ & $2.0 \times 10^{-5}$ & $2.5 \times 10^{-2}$ & $4.6 \times 10^{-4}$ \\
Wave equation & $1.1 \times 10 ^{-3}$ & $5.6 \times 10^{-4}$ & $2.5 \times 10^{-2}$ & $2.9 \times 10^{-4}$ \\
Lotka–Volterra ($\alpha = 0.2$)& $2.5 \times 10^{-3}$ & $2.5 \times 10^{-3}$ & $2.5 \times 10^{-2}$ & $5.3 \times 10^{-3}$ \\
\bottomrule
\end{tabular}%
}
\caption{$\mathrm{MAE}$ of PTL-PINNs, Regular TL, Radau and RK45 for various equations. PTL-PINNs models achieve accuracy of order $10^{-3}$ and $10^{-4}$, which is comparable to RK45 and Radau.}
\label{tab:accuracy_comparisons}
\end{table}

\begin{figure}[H]
	\centering
\includegraphics[width=1.0\linewidth]{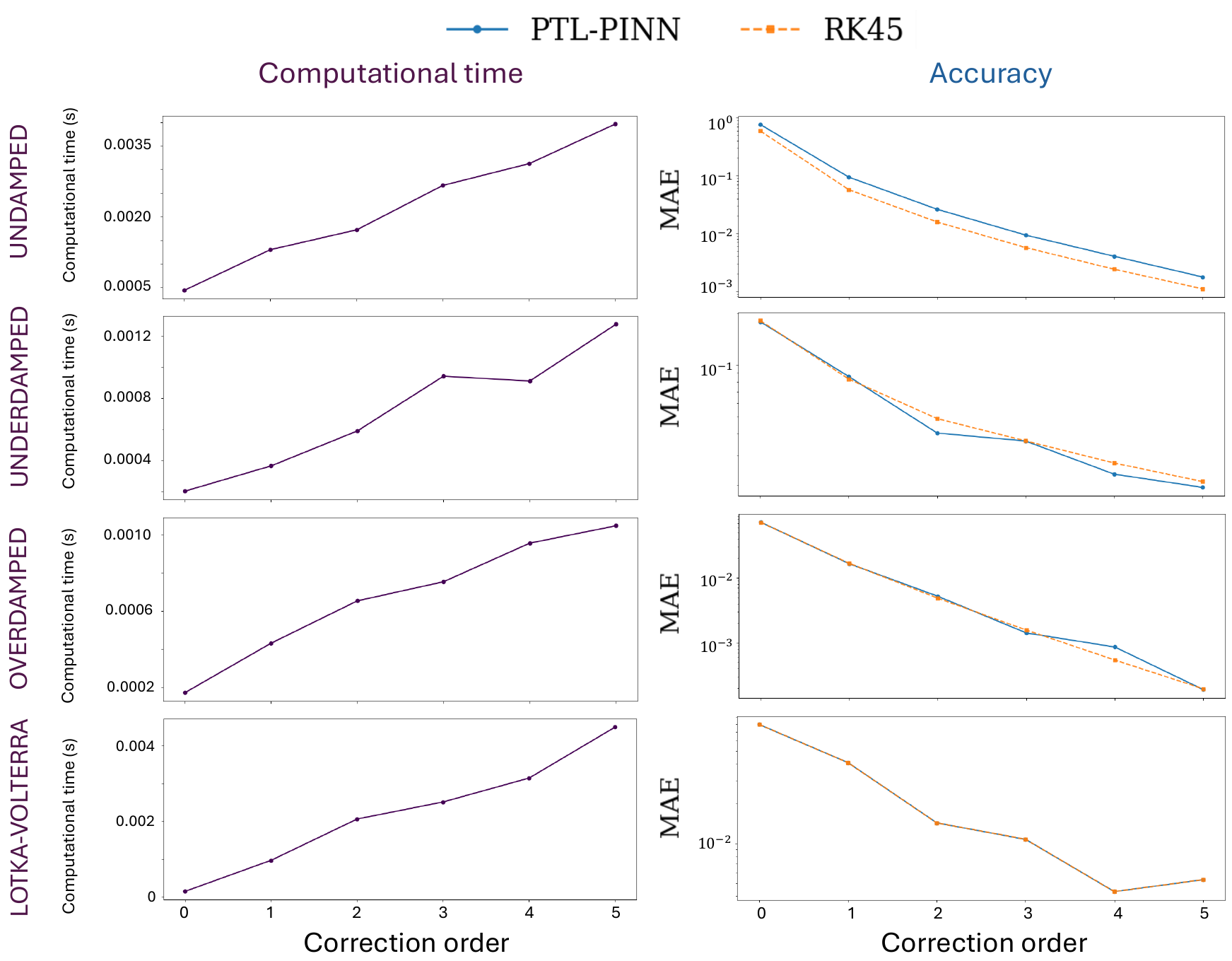}

\caption{Computational time (left panel) and accuracy (right panel) of PTL-PINNs across correction orders for different ODEs and regimes. The computational time increases linearly exponentially with the number of correction terms.}
\label{fig:time_accuracy_vs_order}
\end{figure}

To justify our choice of five correction terms, we examine how the solve time and accuracy vary with the number of corrections when using PTL-PINN-no-invert for the previously considered ODEs, this time including forcing for the damped ODEs. The results are shown in Fig. \ref{fig:time_accuracy_vs_order}. The left plots show the average computational time over twenty-five iterations for each order, revealing a nearly linear increase with the perturbation order. The right plots show that the logarithm of the mean absolute error scales linearly with the number of perturbation terms. This is to be expected since the nth-order terms are scaled by $\varepsilon^n$. In these right plots, we show the theoretical expected accuracy in a dashed line, which was obtained by solving the perturbative system using RK45 with tolerances of $10^{-10}$. We observe that PTL-PINN accuracy across orders follows the theoretical RK45 accuracy across perturbation order. This demonstrates that PTL-PINNs successfully solve the different equations in the perturbative system with high accuracy.

\begin{figure}[H]
	\centering
\includegraphics[width=1.0\linewidth]{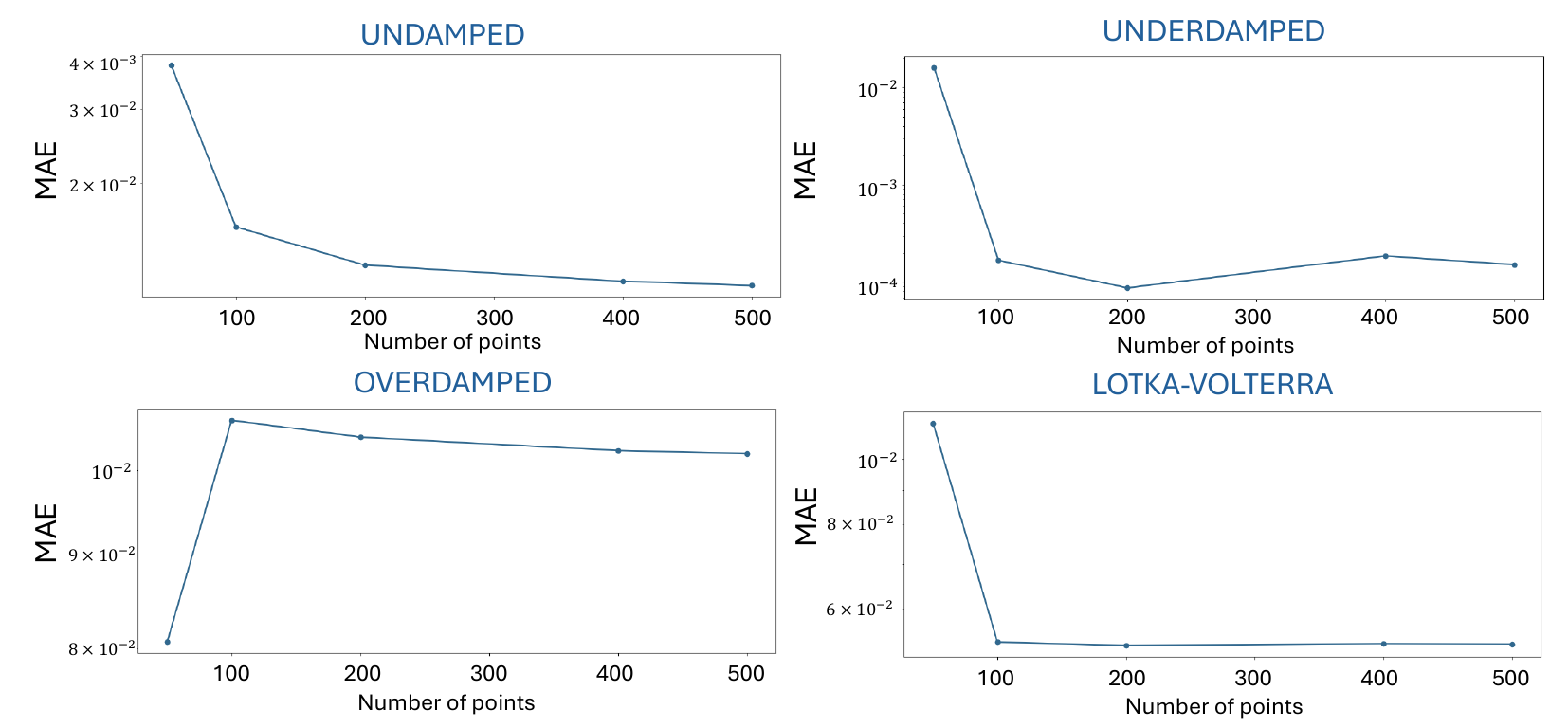}

\caption{Mean absolute error of the PTL-PINN solution vs. number of inference points $N$ for various ODEs. The inference points take values in \{0, 100, 200, 400, 500\}. Apart from $N = 50$, the mean absolute error remains close to constant for different $N$.}
\label{fig:variation_number_points}
\end{figure}

Moreover, in Table \ref{tab:ptl_time_vs_classical_zeta} and \ref{tab:accuracy_comparisons}, we considered $N = 150$ points for ODEs and $N = 50 \times 50$ for PDEs with PTL-PINNs, which for the time interval considered in ODEs, $t \in (0, 10)$, returns a solution evaluated uniformly at points in this interval with time step $\approx 0.067$. The computational time of RK45 and Radau is not significantly influenced by the number of points chosen, since it depends instead on the absolute and relative tolerances considered. However, the more points are considered, the longer PTL-PINNs will take to solve the equation since $H(x,t)$ and the other matrices multiplied will be larger. In Fig. \ref{fig:variation_number_points}, we show that the PTL-PINNs accuracy for ODEs is not affected by the number of points chosen. However, it is important to note that the runtime of this method scales directly with $N$, so $N$ must be chosen to balance computational cost and the desired resolution of the final solution.

The ODEs considered are unforced and the same as the ones detailed for Table \ref{tab:accuracy_comparisons}. 

\section{Perturbation Limitations and Solutions}
\label{sec:perturbation_limitations}

\subsection{Standard Perturbation Method Failure Modes}

\begin{figure}[H]
	\centering
\includegraphics[width=1.0\linewidth]{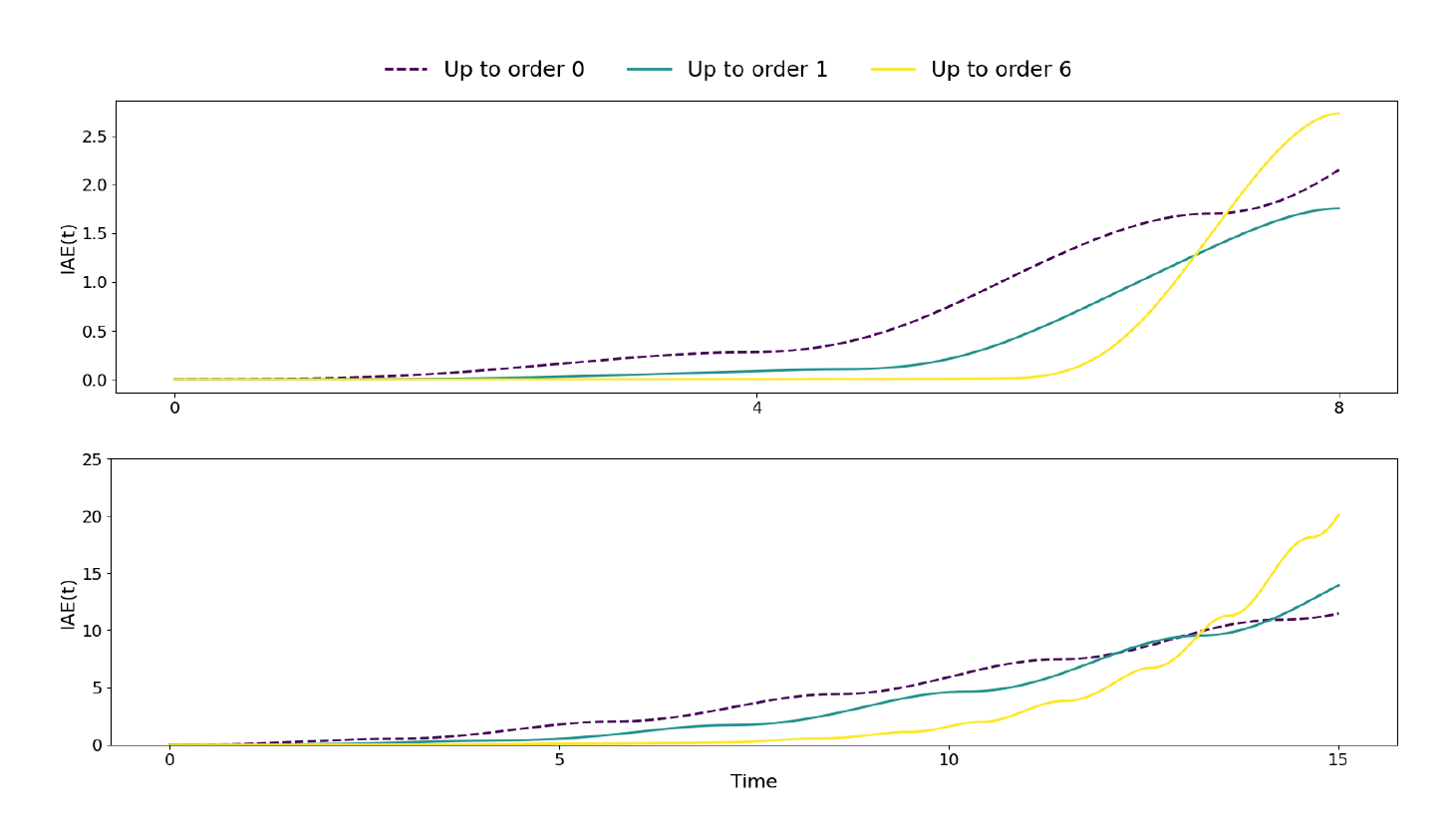}

\caption{Integrated absolute error $\mathrm{IAE}(t)$ of PTL-PINN solutions obtained using standard perturbation theory for
(a) a forced underdamped oscillator and
(b) an unforced undamped oscillator,
as progressively higher-order perturbation corrections are included.
Results are shown for the zeroth-order solution, the first-order correction and the sixth-order correction.
In both cases, higher-order perturbative expansions can lead to increased long-time error due to resonant or near-resonant effects inherent to the standard perturbation method.}
\label{fig:proof_resonance}
\end{figure}

In this subsection, we examine the factors that influence the accuracy of perturbation theory approximations. Fig. \ref{fig:proof_resonance} shows the integrated absolute error, $\mathrm{IAE}(t)$, of the PTL-PINN solutions obtained using the standard perturbation method for both a forced underdamped oscillator and an unforced undamped oscillator, as increasingly higher-order corrections are included. At longer time horizons, we observe that perturbative approximations including additional correction terms can exhibit larger errors.

Naively, one might expect that adding more correction terms to the perturbative series would always improve accuracy. However, the standard perturbation method for the undamped oscillator introduces terms of the form of $t^n \sin(t)$, which correspond to the resonant, non-physical terms discussed in Section \ref{sec:perturbation_theory}. These terms artificially increase the oscillator’s energy and grow with time, which explains the increased error. 

In the underdamped case, there is no pure resonance since the homogeneous solution is not purely oscillatory, taking the functional form of $e^{-\omega_0 \zeta t} \cos\left(\omega_0 \sqrt{1 - \zeta^2} t + \phi\right)$. However, the forcing terms in the perturbative system can have frequencies close to the homogeneous solution, leading to near-resonance effects that degrade accuracy. Such near-resonant behavior arises only in specific regions of the underdamped regime.

Fig. \ref{fig:proof_resonance} (a) illustrates this behavior for an oscillator with very low damping ($\zeta = 0.25$) under external forcing. Due to small damping, the frequency of the harmonics in the homogeneous solution, which is $\sqrt{1 - \zeta^2}$, approaches unity, which coincides with the frequency of the forcing at zeroth-order. Moreover, the exponential decay term, $e^{-\omega_0 \zeta t}$, does not dominate the dynamics over the time interval considered. When the forcing is removed or the damping is increased, the standard perturbation method achieves high accuracy for the same parameters.

Nonetheless, higher-order corrections can improve accuracy in the standard perturbation method over short time scales and for sufficiently small initial conditions,
since the non-physical terms initially have a small amplitude.
The impact of these terms further diminishes as the nonlinearity coefficient $\varepsilon$ decreases,
because $\varepsilon$ scales the magnitude of the perturbative corrections and controls how close the nonlinear frequency remains to the linear one.

These observations suggest that standard perturbation theory can remain effective when combined with early truncation of the series. At the same time, they motivate the use of alternative perturbation approaches, such as the Lindstedt--Poincar\'e method,
when long-time accuracy or oscillatory fidelity is required.

\begin{figure}[H]
	\centering
\includegraphics[width=1.0\linewidth]{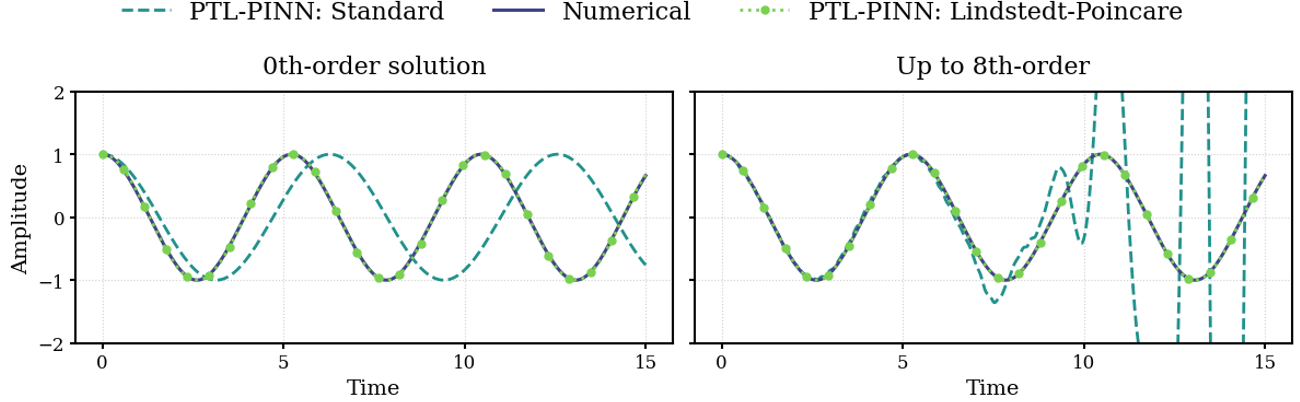}

\caption{PTL-PINN solutions of the unforced and undamped oscillator against an RK45 baseline. The PTL-PINN employing the Lindstedt–Poincaré method accurately recovers the correct solution. The right panel shows PTL-PINN results using only the zeroth-order correction, which has lower error than the left panel results, which consider eight correction terms.}
\label{fig:standard_vs_LPM_undamped}
\end{figure}

In Fig. \ref{fig:standard_vs_LPM_undamped}, we compare the PTL-PINN solution of the unforced undamped oscillator obtained using the standard perturbation and Lindstedt--Poincar\'e methods against the RK45 numerical reference. We observe that the Lindstedt--Poincar\'e formulation successfully removes the resonant terms of the form $t^n \sin(t)$, which were responsible for the monotonic growth of the integrated absolute error with time in Fig. \ref{fig:proof_resonance} (b).

\subsection{Lindstedt--Poincar\'e method with forcing}

Throughout this manuscript, we have primarily considered unforced oscillators when applying PTL-PINNs in combination with the Lindstedt--Poincar\'e method. In this subsection, we show that in the presence of external forcing, the zeroth-order Lindstedt--Poincar\'e solution requires an initial approximation, since the frequency correction series has not yet been determined. As a result, for forced systems, improved accuracy can be achieved by applying the method in multiple passes.

In the first pass, at zeroth-order, we approximate the forcing-phase using $t / \omega  \approx t / \omega_0$. Using this approximation, the perturbation system is solved within the PTL-PINN framework, yielding a first-pass frequency estimate, $\omega^{(1)}$. To improve accuracy, the Lindstedt--Poincar\'e procedure can be rerun, but with the zeroth-order approximation $t / \omega  \approx t / \omega^{(1)}$. 

\begin{figure}[H]
	\centering
\includegraphics[width=0.85\linewidth]{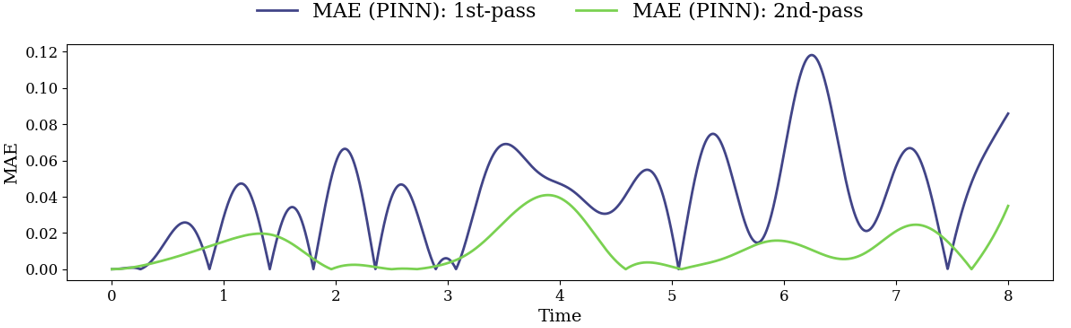}

\caption{PTL-PINN solution mean absolute error for a forced nonlinear oscillator considering the Lindstedt--Poincar\'e method applied once (first pass) and twice (second pass).
In the first pass, the forcing-phase rescaling uses the linear frequency $\omega_0$, yielding a first estimate $\omega^{(1)}$.
In the second pass, the zeroth-order approximation is updated using $\omega^{(1)}$, leading to a refined frequency estimate $\omega^{(2)}$ and improved accuracy.}
\label{fig:error_pass1_vs_pass2}
\end{figure}

In Fig. \ref{fig:error_pass1_vs_pass2}, we illustrate this by comparing the $\mathrm{MAE}$ of the first-pass and second-pass solutions of the Lindstedt--Poincar\'e method for a nonlinearity $0.8 x^3$, external forcing $\cos(t)$ and initial conditions $x(0) = 1$ and $\dot{x}(0) = 0$. The zeroth-order forcing term in the first pass is $\cos(6t / \omega_0) = \cos(6t)$, while in the second pass it becomes $\cos(6t / \omega^{(1)}) \approx \cos(4.74t)$.

\section{Conclusion and future work}
\label{sec:conclusion}

We presented PTL-PINN, a perturbation-guided transfer learning framework, which combines perturbation theory and PINNs. In various benchmark problems, we have successfully demonstrated that PTL-PINNs can accurately and efficiently solve perturbative systems that approximate weakly nonlinear ODEs and PDEs. By using one-shot transfer learning, PTL-PINNs are orders of magnitude faster than gradient-based transfer learning and of comparable speed to RK45 and Radau. This work addresses the fundamental challenges of PTL-PINNs: (1) learning a meaningful latent representation and (2) ensuring the linear system derived from perturbation theory is an accurate approximation.

To address (1), we use the general expected solution of a perturbative system to guide our pretraining. To further improve our results, larger models could be trained on more challenging equations. In addition, regularization strategies can be employed to improve transferability \cite{tarancon2025efficient, wang2025advancing}. To address (2), the perturbation series may need early truncation, alternative perturbation methods should be considered or the method should be avoided altogether for certain parameters. We implemented the Lindstedt--Poincar\'e method for the undamped oscillator and the equilibrium-centered Lotka--Volterra system. Several analytical variants of the Lindstedt--Poincar\'e method \cite{amore2005improved, hu2004comparison} and other perturbation methods, such as the Method of Multiple Scales, will be explored in future work. 

This work lays the groundwork for extending PTL-PINNs to a broader class of nonlinear systems. Extending this framework to higher-dimensional and more complex cases, such as the three-body problem and the Navier-Stokes, which have been studied with perturbation theory \cite{lara2022perturbation, pierson1962perturbation}, will be the focus of future work. PTL-PINNs delivers foundational models that can compute solutions without retraining and with the time complexity of matrix--vector multiplication, broadening the applicability of PINNs to real-life scenarios where computational cost is a limitation.

\section*{Declaration of competing interest}
The authors declare that they have no known competing financial interests or personal relationships that could have
appeared to influence the work reported in this paper.

\section*{Data availability}
The code has been made publicly available on \url{https://github.com/esemsc-dda24/PTL-PINNs}.

\section*{Declaration of AI-assisted technologies}

During the preparation of this work, the authors used ChatGPT (OpenAI GPT-5) to assist with minor language refinement. The tool was not used to generate scientific ideas, analyses, results or interpretations. After using this tool, the authors critically reviewed and verified any suggestions. The authors take full responsibility for the content of the published article.

\section*{CRediT authorship contribution statement}

\textbf{Duarte Alexandrino:} Conceptualization, Methodology, Software, Writing - original draft, Writing
- review \& editing. \textbf{Ben Moseley:} Supervision, Writing
- review \& editing. \textbf{Pavlos Protopapas:} Conceptualization, Supervision, Writing
- review \& editing

\bibliographystyle{elsarticle-num}
\bibliography{bibliography}         

\appendix
\section{Lindstedt--Poincar\'e method}

\subsection{Undamped Oscillator Derivation}
\label{sec:LPM_derivation}

This subsection presents the derivation of the Lindstedt--Poincar\'e method for the undamped oscillator with monomial nonlinearity. A nonlinear forced undamped oscillator is generally described by equation \ref{general_nonlinear_oscillator} for $\zeta = 0$. In the Lindstedt--Poincar\'e method, we perform the variable change $\tau = \omega t$ for the nonlinear equation. Substituting the series expansions for $x$ and $\omega$, each term in the rescaled nonlinear equation can be expressed up to nth-order as follows:
\begin{align*}
  \text{1)} \quad & \omega^{2} \,\ddot{x} 
  = \omega_0^2\,\ddot{x}_0 
  + \cdots 
  + \left( \sum_{k=0}^{n} \ddot{x}_{n-k}\,\sum_{i=0}^{k} \omega_i\,\omega_{k-i} \right)\,\varepsilon^{n} \\
  \text{2)} \quad & \omega^2_0 x = \omega^2_0 x_0 + \cdots + \omega^2_0 x_n \varepsilon^n \\
  \text{3)} \quad & \varepsilon x^q = \epsilon x^q_0 + \cdots + \varepsilon^n \sum_{\substack{k_0 + k_1 + \cdots + k_p = q \\ \sum_{i=0}^{p} i\,k_i = n-1}}
\frac{q!}{k_0!\,k_1!\cdots k_p!}\;
\prod_{i=0}^{p} x_i^{k_i}  \\[1mm]   
  \text{4)} \quad & \sum^{m}_{k = 0} \Gamma_k \cos\left(\Omega_k \frac{\tau}{\omega}\right) \approx \sum^{m}_{k = 0} \Gamma_k \cos\left(\Omega_k \frac{\tau}{\omega_0}\right)
\end{align*}

From the expressions above, we observe that \((1)\) can be decomposed into three terms: one depending on \(\omega_n\), another on \(\ddot{x}_n\) and a final term involving only quantities of order lower than \(n\):

\begin{align*}
  1) \quad \omega^{2}\,\ddot{x} 
  \;=\; \omega_0^2\,\ddot{x}_0 
  \;+\; \cdots \;+\;\Bigl[\,2\,\omega_0\,\omega_n\,\ddot{x}_0 
    \;+\; \ddot{x}_n\,\omega_0^2 
    \;+\; A\bigl(\ddot{x}_0,\,\ddot{x}_{\,n-1},\,\omega_0,\,\ldots,\,\omega_{\,n-1}\bigr)\Bigr]\,
    \varepsilon^n
\end{align*}

Combining all the expressions above, we obtain the general form of the nth-order differential equation for $n > 0$ can be described by equations \ref{eq:general_x_n_undamped} and \ref{eq:specific_f_term}. The left side of equation \ref{eq:general_x_n_undamped} can be separated into a term that depends on $\omega_n$ and another that depends on previously computed quantities from lower orders.

\begin{equation}
    \frac{d^2u}{dt^2} + u_n = \mathcal{F}(u_0, \cdots, u_{n-1}, \omega_0, \cdots, \omega_{n-1}) - 2\frac{\omega_n}{\omega_0} \frac{d^2 u_0}{dt^2}
    \label{eq:general_x_n_undamped}
\end{equation}

\begin{multline}\label{eq:specific_f_term}
\mathcal{F}(x_0,\dots,x_{n-1},\omega_0,\dots,\omega_{n-1}) =
\sum_{\substack{k_0+\cdots+k_p=q\\ \sum_{i=0}^{p} i\,k_i = n-1}}
-\frac{q!}{\omega_0^{2}\,k_0!\,k_1!\cdots k_p!}\,
\prod_{i=0}^{p} x_i^{k_i} \\
- \frac{1}{\omega_0^{2}}\,
A\bigl(\ddot{x}_0,\dots,\ddot{x}_{\,n-1},\omega_0,\dots,\omega_{\,n-1}\bigr)
\end{multline}

To enforce the solvability condition, we calculate the integral of the forcing term in equation \ref{eq:general_x_n_undamped} multiplied by the homogeneous solution of the system. Here, it is sufficient to project the forcing only onto \( \cos \tau \), as the Lindstedt--Poincar\'e method guarantees the elimination of \( \sin \tau \) components by imposing the condition \( \dot{x} = 0 \).

\begin{equation}
    \int^{2 \pi}_{0} \left( \mathcal{F}(x_0, \cdots, x_{n-1}, \omega_0, \cdots, \omega_{n-1}) - 2\frac{\omega_n}{\omega_0} \ddot{x}_0 \right) \cdot \cos \tau = 0
\end{equation}

After isolating $\omega_n$, we obtain the final expression for calculating the frequency correction at each order:

\begin{equation}
    \omega_n = \frac{\int^{2\pi}_0 \mathcal{F}\bigl(u_0,\, \cdots, u_{n-1},\,\omega_0,\,\ldots,\,\omega_{\,n-1}\bigr) \cdot \cos t \: dt}{\int^{2\pi}_0 - \frac{2 \ddot{x}_0}{\omega_0} \; \cos t \; dt}
    \label{eq:w_n_integral}
\end{equation}

\subsection{Equilibrium-centered Lotka--Volterra Derivation}
\label{sec:LK_LPM}

In this subsection, we follow Grozdanovski's weakly nonlinear formulation of the Lotka--Volterra system \cite{grozdanovski2007approximating}. Section \ref{sec:intro_lotka_volterra} introduced this formulation, where $\eta$ and $\xi$ are defined as corrections to the non-trivial critical point ($1,1$), following $x(t) = 1 + \varepsilon \xi(t)$ and $y(t) = 1 + \varepsilon \eta(t)$. Inserting these definitions into \ref{eq:LV}, we obtain the following system:

\begin{equation}
\left\{
\begin{aligned}
    \frac{d \xi}{dt} &= -\eta(t)\big(1 + \varepsilon\, \xi(t)\big), \\[6pt]
    \frac{d \eta}{dt} &= \alpha \xi(t)\big(1 + \varepsilon \eta(t)),
\end{aligned}
\right.
\label{eq:lotka_volterra_system}
\end{equation}

Using the Lindstedt--Poincar\'e method, we introduce the new time variable $\tau = \omega t$, which rescales time and leads to the new system:

\begin{equation}
\left\{
\begin{aligned}
    \omega \frac{d \xi}{d \tau} &= -\eta(\tau)\big(1 + \varepsilon\, \xi(\tau)\big), \\[6pt]
    \omega\frac{d \eta}{d \tau} &= \alpha \xi(\tau)\big(1 + \varepsilon \eta(\tau)),
\end{aligned}
\right.
\label{eq:lotka_volterra_system_rescaled}
\end{equation}

Following the usual perturbation approach, we expand all quantities as power series in the small parameter $\varepsilon$:

\begin{align*}
\xi(\tau) &= \xi_0 + \varepsilon \xi_1 + \varepsilon^2 \xi_2 + \cdots \\[2mm]
\eta(\tau) &= \eta_0 + \varepsilon \eta_1 + \varepsilon^2 \eta_2 + \cdots \\[2mm]
\omega &= \sqrt{\alpha} + \varepsilon \omega_1 + \varepsilon^2 \omega_2 + \cdots
\end{align*}

By inserting these expressions into \ref{eq:lotka_volterra_system_rescaled}, we obtain that the general formula for the nth-order correction is:

\begin{equation}
\left\{
\begin{aligned}
    \omega_0 \frac{d \xi_n}{d \tau} + \eta_n &= - \omega_n \frac{d \xi_0}{d\tau} - \sum^{n - 1}_{i = 1} \omega_i \frac{d \xi_{n - i}}{d \tau} - \sum^{n - 1}_{i = 0} \xi_i \eta_{n -1 - i}
    \\[6pt]
    \omega_0 \frac{d \eta_n}{d \tau} - \alpha \xi_n &= - \omega_n \frac{d \eta_0}{d \tau}  - \sum^{n - 1}_{i = 1} \omega_i \frac{d \eta_{n - i}}{d \tau} + \alpha \sum^{n - 1}_{i = 0} \xi_i \eta_{n -1 - i}
\end{aligned}
\right.
\label{eq:lotka_volterra_system_nth_order}
\end{equation}

To avoid resonance, the solvability conditions must be satisfied, which implies
\[
\int_0^{2\pi} F_{\xi_n}(t)\,\eta_0(t)\,dt = 0, 
\qquad
\int_0^{2\pi} F_{\eta_n}(t)\,\xi_0(t)\,dt = 0
\]
where \( F_{\xi_n} \) and \( F_{\eta_n} \) denote the forcing terms 
in the equations for \( \xi_n \) and \( \eta_n \), respectively. When one of these conditions is satisfied, so is the other. We choose to project $\eta_0$ and after algebraic manipulation of the forcing function, which is represented in \ref{eq:lotka_volterra_system_nth_order}, we find that $\omega_n$ is given by:

\begin{equation}
    \omega_n = \frac{\int \big(  \sum^{n -1}_{i = 1} \omega_i \frac{d \xi_{n-i}}{d \tau} + \sum^{n -1}_{i = 0} \xi_i \eta_{n -1-i} \big) \cdot \eta_0 \, d \tau}{\int^{2 \pi}_0 - \frac{d \xi_0}{d \tau} \cdot \eta_0 d \tau}
\end{equation}

Since we have an expression for $\omega_n$, we can follow the same approach that was outlined for the undamped oscillator to solve the Lindstedt--Poincar\'e system that approximates the equilibrium-centered Lotka--Volterra system.

\section{Pretraining configurations}
\label{sec:training_model}

In this section, we describe the pretraining configurations of the PTL-PINNs models used to obtain the previous results. 

\subsection{ODEs}

For each damping regime, we present the general expected forcing function in the perturbative system and use it to guide the choice of forcing functions in pretraining. Table \ref{tab:undamped_train} shows the configurations of each train head for the undamped model and the equilibrium-centered Lotka--Volterra system. For these equations, the general functional form of the terms present in the forcing of the perturbative system is given by equation \ref{eq:forcing_undamped_lkv}. By expressing the undamped oscillator as a system of two first-order ODEs, it takes the same matrix form as the Lotka--Volterra system. Since both systems admit solutions of the same functional form, we use the same model to solve both equations.

\begin{equation}
    \sum^{m}_{k = 0} \Gamma_k \cos(\Omega_k \,t), \quad \Gamma_k = \{ a_i\}^m_{i = 1} \quad a_i \sim \mathcal{U} \left( 0, \frac{2}{m} \right)
    \label{eq:forcing_undamped_lkv}
\end{equation}

\begin{table}[H]
\centering
\caption{Pretraining configurations of the undamped and Lotka--Volterra PTL-PINN.}
\renewcommand{\arraystretch}{1.1}
\setlength{\tabcolsep}{6pt}
\begin{tabular}{c c c c c}
\toprule
\textbf{Head} & $\Omega_k$ & $\Gamma_k$ & $(x(0),\,\dot{x}(0))$ & $\omega_0$ \\
\midrule
0 & $\emptyset$ & $\emptyset$ & $(1,\,0)$ & 1 \\
1 & $\emptyset$ & $\emptyset$ & $(1.5,\,0)$ & 1 \\
2 & $\emptyset$ & $\emptyset$ & $(0.5,\,0)$ & 1 \\
3 & $\{3\}$ & $\{1\}$ & $(1.2,\,0)$ & 1 \\
4 & $\{3\}$ & $\{1\}$ & $(0.5,\,0)$ & 1 \\
5 & $\{6\}$ & $\{1\}$ & $(1,\,0)$ & 1 \\
6 & $\{3,\,12\}$ & $\mathcal{U}(0,1)$ & $(0,\,0)$ & 1 \\
7 & $\{2,\,4,\,5\}$ & $\mathcal{U}(0,2/3)$ & $(0,\,0)$ & 1 \\
8 & $\{2,\,3,\,4,\,5,\,6\}$ & $\mathcal{U}(0,2/5)$ & $(0,\,0)$ & 1 \\
9 & $\{3,\,6,\,9,\,12,\,15\}$ & $\mathcal{U}(0,2/5)$ & $(0,\,0)$ & 1 \\
10 & $\{6,\,9,\,18,\,21\}$ & $\mathcal{U}(0,2/4)$ & $(0,\,0)$ & 1 \\
11 & $\{3,\,6,\,9,\,12,\,15,\,18,\,21,\,24\}$ & $\mathcal{U}(0,2/8)$ & $(0,\,0)$ & 1 \\
\bottomrule
\end{tabular}
\label{tab:undamped_train}
\end{table}

Table \ref{tab:underdamped_train} shows the configurations of each train head for the underdamped model. In this damping regime, the general functional form of the terms present in the forcing of the perturbative system is given by:

\begin{equation}
    \sum^{m}_{k = 0} \Gamma_k  \,e^{-\mu \zeta \omega_0 \, t} \cos(\Omega_k\, \sqrt{1 - \zeta^2} \,t + \phi), \quad \Gamma_k = \left\{ a_i \right\}_{i=1}^{m} \quad a_i \sim \mathcal{U}\left(0, \frac{2}{m} \right)
\end{equation}

\begin{table}[H]
\centering
\caption{Pretraining configurations of the underdamped PTL-PINN model.}
\renewcommand{\arraystretch}{1.1}
\setlength{\tabcolsep}{6pt}
\begin{tabular}{c c c c c c c}
\toprule
\textbf{Head} & $\zeta$ & $\Omega_k$ & $\Gamma_k$ & $(x(0),\,\dot{x}(0))$ & $\omega_0$ & $\mu$ \\
\midrule
0  & 0.05 & $\emptyset$ & $\emptyset$ & $(1,\,0)$   & 1 & 0 \\
1  & 0.10 & $\emptyset$ & $\emptyset$ & $(1,\,0)$   & 1 & 0 \\
2  & 0.20 & $\emptyset$ & $\emptyset$ & $(1,\,0)$   & 1 & 0 \\
3  & 0.05 & $\{3\}$ & $\{1\}$ & $(1.2,\,0)$ & 1 & 0 \\
4  & 0.10 & $\{3\}$ & $\left\{-\frac{1}{4}\right\}$ & $(0.5,\,0)$ & 1 & 0 \\
5  & 0.10 & $\{6\}$ & $\{1\}$ & $(1,\,0)$ & 1 & 0 \\
6  & 0.50 & $\{3,\,12\}$ & $\left\{-\frac{1}{2},\,-\frac{1}{2}\right\}$ & $(0,\,0)$ & 1 & 0 \\
7  & 0.05 & $\{2,\,4,\,5\}$ & $\mathcal{U}(0,\,2/3)$ & $(0,\,0)$ & 1 & 0 \\
8  & 0.20 & $\{2,\,3,\,4,\,5,\,6\}$ & $\mathcal{U}(0,\,2/5)$ & $(0,\,0)$ & 1 & 0 \\
9  & 0.40 & $\{3,\,6,\,9,\,12,\,15\}$ & $\mathcal{U}(0,\,2/5)$ & $(0,\,0)$ & 1 & 0 \\
10 & 0.10 & $\{6,\,9,\,18,\,21\}$ & $\mathcal{U}(0,\,2/4)$ & $(0,\,0)$ & 1 & 0 \\
11 & 0.05 & $\{3,\,6,\,9,\,12,\,15,\,18,\,21,\,24\}$ & $\mathcal{U}(0,\,2/8)$ & $(0,\,0)$ & 1 & 0 \\
\bottomrule
\end{tabular}
\label{tab:underdamped_train}
\end{table}

Table \ref{tab:overdamped_train} shows the configurations of each train head for the overdamped model. In this damping regime, the general functional form of the terms present in the forcing of the perturbative system is given by:

\begin{equation}
   \Gamma  (\,e^{- \mu_1 \lambda_1 t} + e^{- \mu_2 \lambda_1 \lambda_2t} + e^{- \mu_3 \lambda_1 \lambda_2 t} + e^{-\mu_4 \lambda_1 \lambda_2 t}) \cos(\Omega t + \phi), \quad \Gamma \sim \mathcal{U}\left(0, 2 \right)
\end{equation}

\begin{table}[H]
\centering
\caption{Pretraining configurations of the overdamped PTL\text{-}PINN model.}
\renewcommand{\arraystretch}{1.15}
\setlength{\tabcolsep}{6pt}
\small
\begin{tabular}{c c c c c c c}
\toprule
\textbf{Head} & $\zeta$ & $\boldsymbol{\mu}=(\mu_1,\mu_2,\mu_3,\mu_4)$ & $(x(0),\,\dot{x}(0))$ & $\omega_0$ & $\Omega_k$ & $\Gamma_k$ \\
\midrule
1  & 5  & $(0,0,0,0)$ & $(2.0,\,0)$   & 1 & $\{1\}$ & $\mathcal{U}(0,\,2)$ \\
2  & 10 & $(0,0,0,0)$ & $(1.0,\,0)$   & 1 & $\{1\}$ & $\mathcal{U}(0,\,2)$\\
3  & 20 & $(0,0,0,0)$ & $(1.0,\,0)$   & 1 & $\{1\}$ & $\mathcal{U}(0,\,2)$ \\
4  & 30 & $(0,0,0,0)$ & $(2.0,\,0)$   & 1 & $\{1\}$ & $\mathcal{U}(0,\,2)$ \\
5  & 40 & $(0,0,0,0)$ & $(2.0,\,0)$   & 1 & $\{1\}$ & $\mathcal{U}(0,\,2)$ \\
6  & 50 & $(0,0,0,0)$ & $(1.0,\,0)$   & 1 & $\{1\}$ & $\mathcal{U}(0,\,2)$ \\
7  & 60 & $(0,0,0,0)$ & $(1.0,\,0)$   & 1 & $\{1\}$ & $\mathcal{U}(0,\,2)$ \\
8  & 5  & $(1,0,0,1)$ & $(0.0,\,0)$   & 1 & $\{1\}$ & $\mathcal{U}(0,\,2)$ \\
9  & 30 & $(3,0,0,3)$ & $(0.0,\,0)$   & 1 & $\{1\}$ & $\mathcal{U}(0,\,2)$ \\
10 & 60 & $(3,0,0,3)$ & $(0.0,\,0)$   & 1 & $\{1\}$ & $\mathcal{U}(0,\,2)$ \\
11 & 10 & $(3,2,1,3)$ & $(1.5,\,0)$   & 1 & $\{1\}$ & $\mathcal{U}(0,\,2)$ \\
12 & 30 & $(3,1,2,3)$ & $(1.0,\,0)$   & 1 & $\{1\}$ & $\mathcal{U}(0,\,2)$ \\
13 & 20 & $(3,1,2,3)$ & $(1.0,\,0)$   & 1 & $\{1\}$ & $\mathcal{U}(0,\,2)$ \\
14 & 40 & $(3,0,0,3)$ & $(2.0,\,0)$   & 1 & $\{3\}$ & $\mathcal{U}(0,\,2)$ \\
15 & 20 & $(3,0,0,3)$ & $(1.0,\,0)$   & 1 & $\{5\}$ & $\mathcal{U}(0,\,2)$ \\
\bottomrule
\end{tabular}
\label{tab:overdamped_train}
\end{table}

\subsection{PDEs}
\label{sec:training_config_pdes}

Each linear equation of the perturbative system for the KPP--Fisher is a forced 1D Heat equation. The zeroth-order solution for the initial condition $u(x, 0) = \sin (\pi x/ L)$ can easily be obtained from separation of variables to be:

\begin{equation}
u_0 =e^{-D \left( \frac{\pi}{L} \right)^2 t} \, \sin\!\left( \frac{\pi x}{L} \right)
\end{equation}

Table \ref{tab:KPP_Fisher_configurations} presents the pretraining configurations for the KPP--Fisher equation. More generally, we also consider the forcing terms of the following type: 

\begin{equation}
    e^{- \alpha \lambda_n t} \sin(n \pi x / L), \quad \lambda_n = D \big( \frac{n \pi}{L} \big)^2
\end{equation}

\begin{table}[H]
\centering
\caption{Pretraining configurations of the KPP--Fisher PTL-PINN model.}
\renewcommand{\arraystretch}{1.1}
\setlength{\tabcolsep}{6pt}
\begin{tabular}{c c c c}
\toprule
\textbf{Head} & $D$ & $u(x,0)$ & $f(x,t)$ \\
\midrule
0 & 0.1  & $\sin\!\left(\tfrac{\pi x}{L}\right)$ & 0 \\
1 & 0.05 & $\sin\!\left(\tfrac{\pi x}{L}\right)$ & 0 \\
2 & 0.01 & $\sin\!\left(\tfrac{\pi x}{L}\right)$ & 0 \\
3 & 0.1  & 0 & $u_0(1 - u_0)$ \\
4 & 0.05 & 0 & $u_0(1 - u_0)$ \\
5 & 0.01 & 0 & $u_0(1 - u_0)$ \\
6 & 0.50 & 0 & $n = 1, \, \alpha = 1$ \\
7 & 0.1  & 0 & $n = 2, \, \alpha =2$ \\
8 & 0.05 & 0 & $n = 1, \, \alpha =2$ \\
\bottomrule
\end{tabular}
\label{tab:KPP_Fisher_configurations}
\end{table}

Similarly, each equation in the perturbative system for the Wave equation corresponds to a linear Wave equation. Table \ref{tab:Wave_configurations} shows the configurations used for the pretraining of this PTL-PINN model. The zeroth-order solution for initial condition $u(x, 0) = \sin (\pi x/ L)$ is:

\begin{equation}
    u_0(x,t) = \sin\left(\frac{\pi x}{L}\right)\cos\left(\frac{c \pi t}{L}\right)
\end{equation}

More generally, we can consider forcing functions with arbitrary wave number $k$ and frequency $\omega$:

\begin{equation}
\sin\left(\frac{ k\pi x}{L}\right)\cos\left(\frac{c \omega \pi t}{L}\right), \quad \omega, k \in \mathbb{N} 
\end{equation}

\begin{table}[H]
\centering
\caption{Pretraining configurations of the Wave PTL-PINN model.}
\renewcommand{\arraystretch}{1.1}
\setlength{\tabcolsep}{6pt}
\begin{tabular}{c c c c}
\toprule
\textbf{Head} & $c$ & $u(x,0)$ & $f(x,t)$ \\
\midrule
0 & 1  & $\sin\!\left(\tfrac{\pi x}{L}\right)$ & 0 \\
1 & 0.8 & $\sin\!\left(\tfrac{\pi x}{L}\right)$ & 0 \\
2 & 1.2 & $\sin\!\left(\tfrac{\pi x}{L}\right)$ & 0 \\
3 & 1  & 0 & $u_0(1 - u_0)$ \\
4 & 0.8 & 0 & $u_0(1 - u_0)$ \\
5 & 1.2 & 0 & $u_0(1 - u_0)$ \\
6 & 1 & 0 & $\omega = 1, \, k = 1$ \\
7 & 0.8  & 0 & $\omega = 2, \, k =2$ \\
8 & 1.2 & 0 & $\omega = 1, \, k = 2$ \\
\bottomrule
\end{tabular}
\label{tab:Wave_configurations}
\end{table}

\end{document}